%% file: acl.tex
\pdfoutput=1

\documentclass[11pt]{article}

\usepackage{acl}

\usepackage{times}
\usepackage{latexsym}
\usepackage{graphicx}
\usepackage{multirow}
\usepackage{array}
\usepackage{xcolor}
\usepackage{colortbl}
\usepackage{longtable}
\usepackage[namelimits]{amsmath} 
\usepackage{amssymb} 
\usepackage{amsfonts} 
\usepackage{mathrsfs}
\usepackage{svg}

\usepackage[T1]{fontenc}

\usepackage[utf8]{inputenc}

\usepackage{microtype}

\usepackage{inconsolata}

\input{secs/_custom_things}

\title{Large Language Models are Miscalibrated In-Context Learners}

\author{
Chengzu Li$^{1}$, 
Han Zhou$^{1}$, 
Goran Glavaš$^{2}$,
Anna Korhonen$^{1}$,
Ivan Vulić$^{1}$
\\
$^1$Language Technology Lab, University of Cambridge \\
$^2$Center for Artificial Intelligence and Data Science, University of Würzburg \\
\texttt{\{cl917, hz416, iv250, alk23\}@cam.ac.uk}\\\texttt{goran.glavas@uni-wuerzburg.de}
}

\begin{document}
\maketitle

\input{secs/abstract}

\input{secs/introduction}

\input{secs/related_work}

\input{secs/background}

\input{secs/experiments}

\input{secs/RQ_1}

\input{secs/RQ_2}

\input{secs/7_ablations}

\input{secs/conclusion}

\input{secs/limitations}

\bibliography{anthology,custom}
\newpage
\appendix

\input{appendix/experiment_setup}

\input{appendix/full_results}

\input{appendix/further_ablations}

\input{appendix/prompting_templates}

\end{document}

%% file: secs/_custom_things.tex
\usepackage{xspace}

\newcommand{\rparagraph}[1]{\vspace{1.2mm}\noindent\textbf{#1.}}

\newcommand{\sparagraph}[1]{\vspace{0.0mm}\noindent\textbf{#1.}}

\newcommand{\sparagraphnodot}[1]{\vspace{0.0mm}\noindent\textbf{#1}}

\definecolor{Gray}{gray}{0.92}
\definecolor{racing-green}{rgb}{0.0, 0.8, 0.6}
\definecolor{awesome-red}{rgb}{1.0, 0.13, 0.32}
\newcolumntype{Y}{>{\centering\arraybackslash}X}

\usepackage{booktabs}
\usepackage{makecell}
\usepackage{enumitem}
\usepackage{caption}
\usepackage{subcaption}

%% file: secs/abstract.tex
\begin{abstract}
When adapting ICL with or without fine-tuning, we are curious about whether the instruction-tuned language model is able to achieve well-calibrated results without suffering from the problem of \textit{overconfidence} (i.e., miscalibration) considering its strong instruction following ability, especially in such limited data setups.
In this work, we deliver an in-depth analysis of the behavior across different choices of learning methods from the perspective of both performance and calibration.
Through extensive controlled experiments, we observe that \textit{the miscalibration problem exists across all learning methods in low-resource setups.}
To achieve simultaneous gain for both in-task performance and calibration, we then study the potential of self-ensembling applied at different modeling stages (e.g., variations of in-context examples or variations in prompts or different ensembling strategies) to make the predictions more calibrated and have comparable or even better performance. 
We find that \textit{self-ensembling with max probability produces robust and calibrated predictions}.
Our work reveals \textit{the potential calibration problem of using ICL despite the improvements in task performance} and sheds light on which learning paradigm to choose. 
We also provide \textit{practical guidelines for choosing learning paradigms depending on whether the data has been seen by the model before} and a worthwhile solution via self-ensembling on how to enhance both task performance and calibration of LMs, which we hope could encourage further study.

\end{abstract}

%% file: secs/introduction.tex
\section{Introduction}
Machine learning and NLP have undergone a significant transformation recently, largely propelled by language models (LMs) \citep{Radford2019LanguageMA, brown2020language, Chowdhery2022PaLMSL, OpenAI2023GPT4TR}. 
Among different learning paradigms, Supervised Fine-Tuning (SFT) and In-Context Learning (ICL) have emerged as predominant methodologies \citep{raffel2020exploring, dong2022survey}, demonstrating commendable efficacy across many tasks. 
SFT tunes the model's parameter and effectively \textit{specializes} a (general-purpose) model to specific tasks by learning the knowledge in the training data and optimizing the objective. 
ICL, for each input, leverages the few-shot examples (i.e., the so-called \textit{demonstrations}) to generate predictions without tuning model parameters and treating the model as a `black box'.
Considering the different input format between training with SFT and inference with ICL, \citet{min-etal-2022-metaicl} and \citet{chen-etal-2022-meta} introduce the in-context examples into training phrase, which we call supervised in-context learning (SICL). 
However, when the demonstrations, as a strong inductive bias, get combined with SFT, it has been shown that LMs become more likely to fall into the problem of overconfidence \citep{desai2020calibration, jiang2021can}; the predicted confidence distribution of ICL may be miscalibrated due to the bias in in-context examples \citep{fei-etal-2023-mitigating}. 
Through our extensive experiments, we observe that both paradigms, SFT and ICL, suffer from the problem of miscalibration in low-resource scenarios. 

\input{figs/latex/main_fig}

The important challenges of overconfidence and miscalibration, particularly in scenarios marked by limited data availability, underscore the need for a nuanced understanding of these paradigms. 
These challenges could be more severe in instruction-tuned models considering their strong instruction-following abilities. 
However, most of the previous work \citep{mosbach2023few, sun2023does} only focuses on comparing solely the performance of SFT and ICL on out-of-distribution (OOD) data, targeting general-purpose LMs.  
Here, we instead focus on studying instruction-tuned \textit{task-specialized} language models, where the behavior of different paradigms' in-task performance along with their calibration remains an open research question. 
Therefore, in this work, in addition to the task performance of the models, we are interested in: \textbf{RQ1)} \textbf{how would ICL impact the calibration of LMs?} 
Furthermore, considering the possible issue of overconfidence and miscalibration, we pose and study another crucial research question: \textbf{RQ2)} \textbf{is it possible to ensure both in-task performance and well-calibrated LM behavior at the same time? }
Satisfying both requirements is critical to the application of the model in real-life setups: an applied system should provide both accurate and calibrated predictions to be responsible. 

To address the above challenges, we first investigate the performance and calibration of different model-tuning and ICL methods, along with their interplay, on 7 classification datasets in limited data setups. 
The experiments without ICL demonstrate that  \textit{the vanilla language model is not necessarily calibrated despite of the high in-task performance it achieves}. 
We find empirically that \textit{ICL doesn't help to improve the calibration consistently} across 5 of the 7 classification datasets. 
Our empirical investigations with ICL unveil a \textit{phenomenon between in-task performance and calibration, depending on whether the task dataset has been \textit{seen} by the model before}, which also gains increasing attention over the research community in data contamination \citep{zhu2023clean, deng2023investigating}. 
We further observe that most of the results show relatively high calibration errors, which stands in contrast with the requirements of responsible LMs. 

Based on the findings above, we \textit{propose practical guidelines for choosing different learning paradigms}, which suggests supervised tuning methods are likely to achieve better performance on \textit{unseen} datasets while on \textit{seen} datasets ICL combined with other ‘tweaks’ such as model calibration can be a better choice. 
In response to the possible miscalibration, we explore the application of self-ensembling as a potential solution (shown in Figure \ref{fig:illustration}), inspired by the effect of ensembling of multiple independent models in improving the reliability of the predictions~\citep{ovadia2019can}. 
We incorporate diverse variations in in-context examples and prompts, tailor them to different learning methods, and find that \textit{self-ensembling fortifies the model's calibration by 43\% on average without compromising task performance}.

\rparagraph{Contributions} \textbf{1)} We deliver a comprehensive empirical analysis with different choices of learning methods across a variety of tasks in limited data scenarios, demonstrating that ICL doesn't improve the calibration of the model consistently (\S\ref{sec:main_analysis}). \textbf{2)} We show the relationship between in-task performance and calibration of LMs depending on whether the data has been \textit{seen} by the model and provide practical guidelines for the choice of learning paradigms (\S\ref{sec:guideline}). \textbf{3)} We investigate and justify the feasibility of the self-ensembling pipeline in enhancing both the performance and calibration of LMs (\S\ref{sec:methodology}). We release the code at \url{https://github.com/cambridgeltl/ensembled-sicl}.

%% file: figs/latex/main_fig.tex
\begin{figure}
    \centering
    \includegraphics[width=0.87\columnwidth]{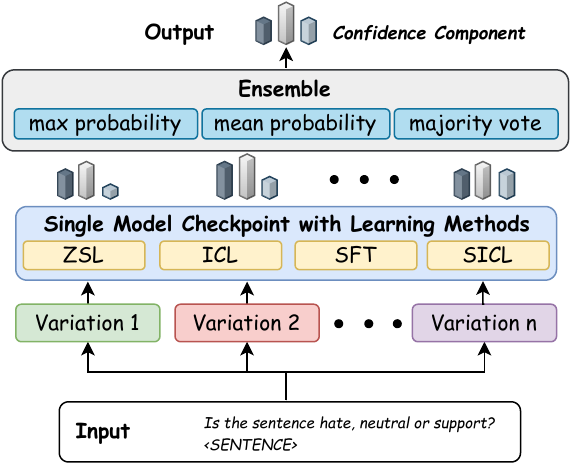}
    \caption{Illustration of the self-ensembled learning methods. We introduce different types of variations to the input and feed them to a single language model. After having the predictions, we run self-ensembling to obtain final predictions and their confidence. }
    \label{fig:illustration}
    \vspace{-2.5mm}
\end{figure}

%% file: secs/related_work.tex
\section{Related Work}

\sparagraph{Learning Paradigms}
Fine-tuning (FT) pretrained LMs has been used as an effective method to adapt them to specific tasks and datasets \citep{devlin-etal-2019-bert, raffel2020exploring}.  With the generation of much larger and more powerful LMs \citep{brown2020language, chung2022scaling}, parameter-efficient fine-tuning (PEFT) \citep{he2021towards, zhou2023autopeft} has been proposed, where the central idea is to tune only a fraction of the model parameters to reduce computation and memory costs. 

Without tuning the model, in-context learning (ICL) has shown great potential, achieving promising performance on various tasks with demonstration examples \citep{brown2020language}. 
Motivated by the positive results of ICL by concatenating multiple in-context (IC) examples to the LM input at inference, \citet{min-etal-2022-metaicl} and \citet{chen-etal-2022-meta} introduce labeled in-context examples to the supervised training process (SICL). This has been further improved and utilized with pretraining \citep{gu-etal-2023-pre, shi2023context} and other training strategies \citep{ye2023context, wei2023symbol}. 

\citet{sun2023does} study task performance and stability with different PEFT methods in addition to ICL, including prompt tuning, and instruction tuning (IT)~\citep{singhal2022large}. 
\citet{duan2023exploring} explore the relationship between ICL and IT, and interpret ICL as implicit IT. \citet{mosbach2023few} compare the generalization ability of ICL and FT on out-of-distribution (OOD) data. \citet{zhou-etal-2023-survival} show the superior performance of hard prompt tuning over standard FT in low-data scenarios. However, previous work has not explored all the learning methods systematically within low-data scenarios and has not investigated them through the joint optics of in-task performance, confidence, and their trade-offs.

\sparagraph{Calibrating LMs}
A well-calibrated LM should be accurate in terms of performance while also producing reliable confidence estimates for their predictions. 
When concatenating different in-context examples with various labels, tokens, and example ordering, ICL would be influenced by the bias in the concatenated examples~\citep{zhao2021calibrate}, and various calibration methods have been proposed to mitigate this issue~\citep{wang2023large, zhou2023batch, liu2024aligning}. 
Moreover, the model itself also contains certain `implicit' label biases due to its (pre)training corpus, which would have an effect on the confidence estimation as well~\citep{fei-etal-2023-mitigating}. 
Furthermore, FT may suffer from miscalibration for both in-distribution and OOD data due to over-parameterization when adapting to the specific data \citep{kong2020calibrated}, and the miscalibration effect, as we investigate in this paper, might be even more pronounced in limited data scenarios.

\sparagraphnodot{Ensembling Model Predictions}
has been used to mitigate the problem of overconfidence and improve the reliability of the final model predictions~\citep{ovadia2019can}. A standard ensembling practice is to train the model with different hyper-parameters or different initialization \citep{wenzel2020hyperparameter, lakshminarayanan2017simple}.
Concerning LMs, \citet{sun2022quantifying} ensemble fine-tuned LMs to quantify the uncertainty with disagreement among different ensembling components. 
\citet{gleave2022uncertainty} and \citet{wang2023lora} ensemble partially tuned LMs, considering the computation resources for training and the storage for saving different LM checkpoints. 
Although the proposed method achieves better and more reliable results, it takes a considerable amount of time, computation, and storage resources to train and save multiple models, which often makes them inapplicable to LMs. 
In contrast to having several tuned models with supervised learning, \citet{yao2023more} demonstrate the feasibility of using self-ensembling in ICL without tuning the model. 
In this work, we explore self-ensembling in the novel context of diverse learning paradigms and low-data setups and show that it is possible to improve model calibration without compromising performance.

%% file: secs/background.tex
\section{Background: Learning Paradigms}
\label{sec:background}
In this paper, we analyze and compare four different learning paradigms relevant to ICL in low-resource scenarios: zero-shot learning  (ZSL),\footnote{In ZSL and later ICL there is no actual 'learning' taking place, and the model simply reacts to the provided prompt, but we have chosen the term ZSL for consistency with previous literature.} in-context learning (ICL), supervised fine-tuning (SFT) and supervised in-context learning (SICL). With classification tasks in focus, we briefly describe each paradigm in what follows. 

\sparagraph{Zero-Shot Learning (ZSL)}
Given the input $x$ and the prompting template $f_{p}$, the prediction $\hat{y}$ from the LM can be represented as 
$\hat{y} = {{\arg\max}}_j\ \mathcal{P}(y_j|f_p(x))$, 
where the parameters of the underlying LM are fixed.
The prompting template $f_p(x)$ includes the task instructions and special symbols which can be replaced by the input $x$. 
We attach the prompting templates for different classification tasks in Appendix~\ref{app:prompt template}. 

\sparagraph{In-Context Learning (ICL)}
Similar to ZSL, instead of only feeding the input $x$ to the model, we first prepend $M$ in-context examples (IC) (also called demonstrations) $[f_p(x_{1}), y_1; ...; f_p(x_{M}), y_M]$ to the input $x$. The examples are retrieved from the pool of examples $\rm R$ following (random or non-random) selection strategy. 
The prediction is then defined as
$\hat{y} = {{\arg\max}}_j\ \mathcal{P}(y_j|[f_p(x_{IC}), y_{IC}], f_p(x))$. 

\sparagraph{Supervised Fine-Tuning (SFT)}
As mentioned, ZSL and ICL are inference-only paradigms treating the LM as a black box. On the other hand, SFT first trains the model on the training set following the input format $f_p(x)$ from ZSL. Note that here we use SFT to refer to instruction-style fine-tuning with a prompting template, which we see as the context. Inference with the tuned model $\mathcal{P'}$ is then conducted in the same way as with ZSL. During training and inference, we can use different prompting templates to create variations in the model input, which we further elaborate on in \S\ref{subsec:variations}. 

\sparagraph{Supervised In-Context Learning (SICL)}
Based on the propositions from \citet{min-etal-2022-metaicl} and \citet{chen-etal-2022-meta}, we can also fine-tune the model to directly optimize the in-context learning objective. 
For each training step, $M$ in-context examples $(x_1, y_1), ..., (x_M, y_M)$ are selected from the pool $\rm R$. We then prepend the selected in-context examples to the input $x$ as before with ICL and use the concatenation as the final model input, and train the model to generate $y$. Inference proceeds in the same way as with ICL, except that we now use the task-tuned model $\mathcal{P'}$.

%% file: secs/experiments.tex
\section{Experimental Setup}
\label{sec:experimental}

\sparagraph{Datasets and Evaluation Metrics}
We consider 7 classification datasets that cover a range of label numbers and scenarios: SST-2, SST-5 \citep{socher-etal-2013-recursive}, RTE \citep{wang2019glue}, ANLI \citep{nie-etal-2020-adversarial}, Measuring Hate Speech corpus \citep{sachdeva-etal-2022-measuring}, Intent Detection from NLU++ \citep{casanueva-etal-2022-nlu} and Manifestos~\citep{Lehmann:2023}. In order to simulate low-data setups, we sub-sample smaller training data from the full data for each dataset. The details of the datasets along with their corresponding evaluation metrics are provided in Appendix~\ref{app:dataset detail}.

\sparagraph{Implementation Details}
Unless noted otherwise, we use Flan-T5\textsubscript{large} \citep{chung2022scaling} as the main model in the experiments. 
Detailed training environments and hyper-parameters are provided in Appendix~\ref{app:environment} and~\ref{app:hyperparameters}. Further, we provide all the prompting templates in Appendix~\ref{app:prompt template main} and~\ref{app:prompt template cycling}. 

\sparagraph{Estimating Calibration}
Beyond task performance of all the possible variants, we estimate the calibration of the model's predictions (as a proxy towards model confidence) by using Expected Calibration Error (ECE)~\citep{guo2017calibration}. It divides the $n$ predicted results based on their confidence into $M$ bins $B_1$ to $B_M$ and then computes a weighted average over the absolute difference between the accuracy ${\rm acc}(B_m)$ and mean confidence ${\rm conf}(B_m)$ of the predictions within each bin. We set $M$ to 10 in this work. 
\begin{equation}
    \widehat{\rm ECE} = \sum_{m=1}^M \frac{|B_m|}{n}|{\rm acc}(B_m)-{\rm conf}(B_m)|
\end{equation}
ECE measures the difference between the model's empirical accuracy and its confidence (predicted probability). The smaller the ECE, the more confident the model prediction would be. We also report the negative log-likelihood (NLL) \citep{hastie01statisticallearning} $-\sum_{i=1}^n{\rm log}(p_i)$ and information entropy (IE) $-\sum_{i=1}^np_i{\rm log}(p_i)$ as supplementary metrics of model's (lack of) confidence \citep{zhang-etal-2025-get}. 
We define well-calibrated results by an ECE threshold below 0.2 to facilitate clearer discussion.

%% file: secs/RQ_1.tex
\section{RQ1: How does ICL Impact Model Calibration?}
In this section, we are interested in how ICL impacts the calibration of the language model.
To answer this question, we experiment with ZSL, ICL, SFT and SICL to find out 1) whether the model is calibrated and 2) the differences in calibration with different learning methods. 
We first present a comprehensive analysis of various learning methods in low-resource settings, detailing model performance and calibration errors in Table \ref{tab:main-results}. Full experimental results are shown in Appendix \ref{sec:full results}.

\input{tables/main_results}

\subsection{Results and Discussions}
\label{sec:main_analysis}

\sparagraph{The language model with ICL is not necessarily calibrated despite of the high in-task performance}
We observe that when testing the model performance with the least inductive bias using ZSL, the model shows relatively high ECEs on SST-2 and RTE despite that it achieves more than 80\% accuracy, and is only calibrated on SST-5 (ECE \textless 0.2).
When we use ICL without tuning the model, we only witness a significant drop of ECE in 2 of 7 datasets. 
By tuning the model with SFT or SICL, the model becomes even less calibrated on SST-2, RTE, SST-5, although it achieves better in-task performance.
Based on this finding, we further look into the interplay between the in-task performance and model calibration.

\input{tables/trustworthiness_results}

\rparagraph{Performance and calibration of learning methods are task-dependent, depending on whether the data has been seen by the model}
We find that learning methods perform differently depending on the datasets and we divide the tasks into different families depending on their observed behavior. ICL demonstrates comparable performance to SFT/SICL on SST-2 and RTE. 
However, tuning on these datasets with SFT/SICL yields increased ECE along with higher NLL and lower IE scores as shown in Table~\ref{tab:trustworthiness results}, but no substantial in-task performance improvement. 
This indicates that the model does not recover the ground truth distribution in the test set while becoming more confident and certain about its predictions, which serves as a sign of miscalibration. 
Conversely, tasks such as intent detection (NLU++), Manifestos, and Hate speech, show noticeable performance enhancement and better calibration with lower ECE by using SFT/SICL.
Nevertheless, despite these task-dependent variations, ECE remains relatively high across all methods except for intent detection, indicating the problem of \textit{miscalibration across all learning methods}.

\rparagraph{ICL can achieve comparable performance with SFT on `seen' data} 
We suspect the divergent behaviors are possibly due to data contamination of FLAN training corpus \citep{longpre2023flan} wherein ZSL and ICL exhibit similar performances with SFT and SICL on training datasets labeled as \textit{seen} (e.g., SST-2, RTE).\footnote{We corroborate findings from other concurrent work on data contamination~\citep{zhu2023clean, deng2023investigating} that also reports the unfair practice of evaluations on seen datasets. } 
To further investigate the performance on \textit{seen} datasets, we apply the Batch Calibration method \citep{zhou2023batch}, as shown in Table~\ref{apptab:main full results} in the Appendix. 
Surprisingly, we find that ICL performs on par or even better than SICL with calibration across all the (possibly) seen data, which reveals the ability of LMs that have been recovered by calibration techniques on these seen tasks. However, for unseen datasets (NLU++, Manifestos, etc.), the performance of ICL, even with the calibration method applied, is not comparable to those of either SICL or SFT. 

\subsection{Practical Guideline on Choosing Learning Methods}
\label{sec:guideline}
Given the comparison of the performances and calibration on different datasets, we suggest that the choice of learning methods should be task-dependent.
The experiments and analysis indicate that unseen datasets obtain better performance and more trustworthy results with supervised tuning methods. 
For the seen datasets, ICL combined with other `tweaks' such as model calibration can be a better choice, since the supervised tuning methods are more likely to make the model over-confident and less trustworthy. 

Within supervised tuning methods, for SFT and SICL, we empirically observe that SICL shows marginally higher performance ($\uparrow1.23$) and lower ECE ($\downarrow0.05$) than SFT on average across unseen datasets in Table \ref{tab:main-results}. 
We believe this is possibly due to \textbf{1)} the knowledge in the IC examples in addition to the training input-label pairs and \textbf{2)} different combinations of in-context examples as a way of data augmentation in low-resource scenarios.

%% file: tables/main_results.tex
\begin{table*}[t]
\centering
\renewcommand\arraystretch{1.05}
\resizebox{\linewidth}{!}{%
\begin{tabular}{l|l|lllllll} 
\Xhline{2\arrayrulewidth}

\textbf{Metrics}                                   & \textbf{Methods}         & \textbf{\textit{SST-2}}                & \textbf{\textit{RTE}}                 & \textbf{\textit{ANLI}}                & \textbf{SST-5}                & \textbf{NLU++}               & \textbf{Manifestos}          & \textbf{Hate Speech}          \\ 
\hline
\multirow{4}{*}{\textbf{Performance}}     & \textbf{ZSL}    & 94.67                        & 86.64                        & 52.30                        & 42.00                        & 29.20                        & 14.50                        & 37.08                         \\
                                          & \textbf{ICL}    & 95.22\textsubscript{0.12}    & 88.45\textsubscript{0}       & 52.17\textsubscript{0.47}    & 37.59\textsubscript{0.23}    & 40.11\textsubscript{0.09}    & 13.01\textsubscript{0.19}    & 40.09\textsubscript{0.08}     \\
                                          & \textbf{SFT}     & 95.61\textsubscript{0.20}    & \textbf{88.81}\textsubscript{0.29}    & 61.63\textsubscript{1.68}    & 46.27\textsubscript{2.09}    & 79.98\textsubscript{0.59}    & 35.76\textsubscript{1.23}    & 58.01\textsubscript{1.01}     \\
                                          & \textbf{SICL} & \textbf{95.63}\textsubscript{0.29}    & 88.57\textsubscript{0.45}    & \textbf{63.90}\textsubscript{0.14}    & \textbf{47.12}\textsubscript{1.93}    & \textbf{80.76}\textsubscript{0.31}    & \textbf{37.55}\textsubscript{1.61}    & \textbf{59.48}\textsubscript{1.79}     \\ 
\hline
\multirow{4}{*}{\textbf{ECE}} & \textbf{ZSL}    & \textbf{0.907}                       & \textbf{0.809}                       & 0.356                       & \textbf{0.142}                       & 0.231                       & 0.432                       & 0.318                        \\
                                          & \textbf{ICL}    & 0.915\textsubscript{0.001} & 0.815\textsubscript{0.003}   & 0.351\textsubscript{0.005} & 0.183\textsubscript{0.002} & 0.129\textsubscript{0.000} & 0.476\textsubscript{0.002} & 0.271\textsubscript{0.002}  \\
                                          & \textbf{SFT}     & 0.941\textsubscript{0.011} & 0.842\textsubscript{0.002} & 0.316\textsubscript{0.023} & 0.403\textsubscript{0.032} & 0.011\textsubscript{0.001} & \textbf{0.201}\textsubscript{0.072} & 0.354\textsubscript{0.036}  \\
                                          & \textbf{SICL} & 0.945\textsubscript{0.013} & 0.876\textsubscript{0.003} & \textbf{0.280}\textsubscript{0.011} & 0.360\textsubscript{0.025} & \textbf{0.002}\textsubscript{0.001} & 0.214\textsubscript{0.038} & \textbf{0.193}\textsubscript{0.113}  \\
\Xhline{2\arrayrulewidth}

\end{tabular}
}
\vspace{-1mm}
\caption{Results for different learning methods across all 7 datasets. We report the average of 3 independent runs with different random seeds; variance is reported in the subscript. Numbers in \textbf{bold} represent the best performance and calibration score per dataset. The datasets `seen' by Flan-T5 at pretraining are labeled in \textit{italic}.}
\label{tab:main-results}
\vspace{-1.5mm}
\end{table*}

%% file: tables/trustworthiness_results.tex
\begin{table}
\centering
\resizebox{\linewidth}{!}{%
\begin{tabular}{ll|cc|cc} 
\Xhline{2\arrayrulewidth}

\multicolumn{2}{c|}{\multirow{2}{*}{\textbf{Evaluation Metrics }}}   & \multicolumn{2}{c|}{\textbf{SST-2 }}  & \multicolumn{2}{c}{\textbf{SST-5 }}        \\
\multicolumn{2}{c|}{}                                                & \textbf{ICL} & \textbf{SICL}       & \textbf{ICL} & \textbf{SICL}            \\ 
\hline
\multirow{3}{*}{\textbf{Calibration }} & \textbf{ECE}            & 0.915       & 0.945                & 0.183       & 0.360                     \\
                                           & \textbf{NLL}            & 0.135       & 0.271                & 1.226       & 2.339                     \\
                                           & \textbf{IE}             & 0.056       & 0.015                & 0.152       & 0.049                     \\ 
\hline
\multicolumn{2}{l|}{\multirow{2}{*}{\textbf{Evaluation Metrics }}}   & \multicolumn{2}{c|}{\textbf{NLU++ }} & \multicolumn{2}{c}{\textbf{Manifestos }}  \\
\multicolumn{2}{l|}{}                                                & \textbf{ICL} & \textbf{SICL}       & \textbf{ICL} & \textbf{SICL}            \\
\hline
\multirow{3}{*}{\textbf{Calibration }} & \textbf{ECE}            & 0.129       & 0.002                & 0.476       & 0.214                     \\
                                           & \textbf{NLL}            & 0.214       & 0.084                & 3.942       & 2.026                     \\
                                           & \textbf{IE}             & 0.142       & 0.002                & 0.101       & 0.145                     \\
\Xhline{2\arrayrulewidth}

\end{tabular}
}
\vspace{-1mm}
\caption{Calibration errors and other uncertainty metrics of different learning methods across tasks (part of). We refer the readers to Appendix \ref{app:full main results} for full results. }
\label{tab:trustworthiness results}
\vspace{-1.5mm}
\end{table}

%% file: secs/RQ_2.tex
\section{RQ2: How to Ensure Performance and Calibration?}
\label{sec:methodology}

\subsection{Self-Ensembling}
So far, we have observed the common miscalibration issues for all learning methods. We then investigate the feasibility of self-ensembling to improve calibration.
There are two points that create possible variations that can be used for ensembling: \textbf{1)} variation in the selection of in-context examples (for ICL and SICL), and \textbf{2)} variation in the chosen prompt (for all the paradigms). Previous work focuses on \textbf{1)} selecting a better combination of in-context examples~\citep{su2022selective} for the model or \textbf{2)} generating an optimal prompting template~\citep{zhou-etal-2023-survival}. 
On the other side, how the variation of multiple demonstration combinations and prompting templates influences the model behavior is still unexplored. Furthermore, we can \textbf{3)} `self-ensemble' the model with different ensembling strategies. We now introduce these variants.

\subsubsection{Variation of Ensembling Components}
\label{subsec:variations}

\sparagraph{Variation of In-Context Examples (\textit{Var-IC})}
For ICL and SICL, IC examples and their ordering $[f_p(x_{IC}), y_{IC}]$ create variations in the model inputs with a fixed template $f_p$, while not impacting the test pair $f_p(x)\sim y$. 
This allows us to create various in-context example combinations as different inputs to a single model and obtain different ensemble components. 

\rparagraph{Variation of Prompting Templates (\textit{Var-Prompt})}
Different prompting templates have shown high variance in task performance~\citep{mishra2021reframing}. 
By changing the wording in the templates $f_p$, we can also create variations even with the same input to the model. 
For each input $x$, we randomly select a prompting template $f_p'$ from a set of available prompting template candidates.  In ICL and SICL, the same template is also applied to the in-context examples, formatting the final input as $[f_p'(x_{1}), y_1; ...; f_p'(x_{M}), y_M, f_p'(x)]$. This makes it applicable not only to ICL and SICL, but also to ZSL and SFT as well. 

\sparagraph{Variation of Both (\textit{Var-Both})}
When we create a set of ensembling components, we can also combine these two variations.

\subsubsection{Self-Ensembling Strategy}
For each variant, we obtain the predicted results $\hat{y}$ and the confidence $\hat{p}$ for each component. 
The next step involves ensembling the predictions over $K$ different components. 
We experiment with three (self-) ensembling strategies to compare their impact on both performance and calibration.

\rparagraph{Majority Vote}
We select the predicted results that have the highest accumulated probability across $K$ variants as the ensembling predictions. 
The accumulated probability $\mathcal{P}_{acc}$ for the predicted label $l_i$ is defined as
$\mathcal{P}_{acc}(\hat{y}=l_i) = \sum_{k=1}^{K}\mathcal{P}_{k}(\hat{y_k}=l_i)\mathbb{I}(\hat{y_k}=l_i)$. 
We pick the variants that have the same prediction as the ensembling prediction and average the probability distribution of the selected components $\mathcal{P}_{ens}(y|x) = \frac{1}{K'}\sum_{k=1}^{K'}\mathcal{P}_k(y|x)$
where $K'$ is the number of selected variants. 

\rparagraph{Mean Probability}
We average the predicted probability distribution of $K$ variants and use the prediction that has the largest probability in the averaged distribution as the ensemble result 
$\hat{y_{ens}} = {\arg\max}_j\ \mathcal{P}_{ens}(y_j|x)$, where $\mathcal{P}_{ens}(y|x)$ is described as $\mathcal{P}_{ens}(y|x) = \frac{1}{K}\sum_{k=1}^{K}\mathcal{P}_k(y|x)$.

\rparagraph{Max Probability}
For each possible value in the output space, we find the maximum probability of the predicted values across $K$ variants and use this as the prediction's probability $\mathcal{P}'(\hat{y} = l_{i}|x)=\max(\mathcal{P}_j(\hat{y} = l_{i}|x), j \in [1, K])$. 
\noindent Because the probability is obtained from different components, the summation of these probabilities is not guaranteed to be 1. Therefore, we apply the normalization on the new probability distribution: $\mathcal{P}_{ens}(y|x) = {\rm Norm}(\mathcal{P}'(y|x))$. The ensemble prediction is determined as the $\hat{y}$ that has the highest probability after the ensembling step.

\subsection{Results and Discussions}

\input{tables/self_ensemble_results}

\rparagraph{Self-ensembling works across learning methods and enhances calibration performance}
In Table~\ref{tab:self ensembling results}, with different learning methods combined with self-ensembling variations in the design, we find that by changing the in-context example combinations or prompting templates, the best performance of self-ensembling outperforms the baseline without any ensembling by 0.79. Even though the performance gains seem marginal, self-ensembling substantially enhances the calibration performance, reducing the mean ECE value by 43\%. 
The empirical results also show that in addition to ICL, SFT, and SICL also benefit from self-ensembling in both performance and calibration scores. 
SFT and SICL exhibit a larger drop in ECE after self-ensembling than ICL. 
We also notice that when self-ensembling over SFT and SICL, the model has lower ECE scores than ICL, but with much better task performance. 
This indicates the efficiency of self-ensembling in making the predictions more trustworthy while maintaining or even improving the task performance. 
It also suggests that \textit{self-ensembling has the potential to mitigate the prominent problem of overconfidence in supervised tuning methods}, as shown by Figure \ref{fig:illustration}.

\rparagraph{Different Variations and Ensembling Strategies}
Our results suggest that with ICL, \textit{Var-IC} yields more improvements than \textit{Var-Prompt}, while the latter shows its efficacy with SFT and SICL. 
This difference stems from inappropriate prompting templates, requiring model tuning to adhere to the prompt structure. 
We also find that combining both variations may not necessarily improve the performance but is helpful in enhancing the trustworthiness empirically. 
Regarding ensemble strategies, we notice that the majority vote improves the performance in general, but struggles to reduce the calibration error. 
Ensembling with max probability consistently produces the most faithful predictions with promising performances. 
This can be explained by the idea that normalizing over the max probabilities in a way smooths the probability distribution, making the max probabilities less extreme and mitigating over-confidence issues.

%% file: tables/self_ensemble_results.tex
\begin{table*}[t!]
\centering
\resizebox{0.95\linewidth}{!}{%
\begin{tabular}{l|ccccc|ccccc} 
\Xhline{2\arrayrulewidth}

\multirow{3}{*}{\textbf{Systems }}       & \multicolumn{10}{c}{\textbf{Manifestos }}                                                                                                                                                                                                                        \\
                                         & \multicolumn{5}{c}{\textbf{Macro F1 }}                                                                                       & \multicolumn{5}{c}{\textbf{ECE }}                                                                                                 \\
                                         & \textbf{Ori.} & \textbf{Max} & \textbf{Mean} & \textbf{Majority} & \multicolumn{1}{c}{$\Delta$}                & \textbf{Ori.} & \textbf{Max} & \textbf{Mean} & \textbf{Majority} & $\Delta$                                         \\ 
\hline
\rowcolor[rgb]{0.900,0.900,0.900} ZSL    & 14.50           &                   &                    &                        & \textcolor[rgb]{0,0.502,0}{$\uparrow0.79$} & 0.432          &                   &                    &                        & \textcolor[rgb]{0,0.502,0}{$\downarrow0.170$}  \\
\textit{+ Var-Prompt}                & 14.50           & 13.69             & 12.93              & \textbf{15.29}                  & \textcolor[rgb]{0,0.502,0}{$\uparrow0.79$} & 0.432          & \textbf{0.262}            & 0.335             & 0.437                 & \textcolor[rgb]{0,0.502,0}{$\downarrow0.170$}  \\
\rowcolor[rgb]{0.900,0.900,0.900} ICL    & 13.01           &                   &                    &                        & \textcolor[rgb]{0,0.502,0}{$\uparrow0.68$} & 0.476          &                   &                    &                        & \textcolor[rgb]{0,0.502,0}{$\downarrow0.283$}  \\
\textit{+ Var-IC}              & 13.01           & 13.50             & 13.46              & \textbf{13.69}                  & \textcolor[rgb]{0,0.502,0}{$\uparrow0.68$} & 0.476          & 0.415            & 0.465             & 0.469                 & \textcolor[rgb]{0,0.502,0}{$\downarrow0.061$}  \\
\textit{+ Var-Prompt}                & 13.01           & 13.25             & 11.67              & 11.53                  & \textcolor[rgb]{0,0.502,0}{$\uparrow0.24$} & 0.476          & 0.268            & 0.366             & 0.472                 & \textcolor[rgb]{0,0.502,0}{$\downarrow0.208$}  \\
\textit{+ Var-Both}                          & 13.01           & 11.19             & 11.50              & 11.21                  & \textcolor{red}{$\downarrow1.51$}          & 0.476          & \textbf{0.193}            & 0.354             & 0.480                 & \textcolor[rgb]{0,0.502,0}{$\downarrow0.283$}  \\
\rowcolor[rgb]{0.900,0.900,0.900} FT     & 35.76           &                   &                    &                        & \textcolor[rgb]{0,0.502,0}{$\uparrow0.73$} & 0.201          &                   &                    &                        & \textcolor[rgb]{0,0.502,0}{$\downarrow0.134$}  \\
\textit{+ Var-Prompt}                & 35.39           & \textbf{36.49}             & 35.66              & 34.91                  & \textcolor[rgb]{0,0.502,0}{$\uparrow1.10$} & 0.144          & \textbf{0.066}            & 0.105             & 0.135                 & \textcolor[rgb]{0,0.502,0}{$\downarrow0.077$}  \\
\rowcolor[rgb]{0.900,0.900,0.900} SupICL & 37.55           &                   &                    &                        & \textcolor[rgb]{0,0.502,0}{$\uparrow0.06$} & 0.214          &                   &                    &                        & \textcolor[rgb]{0,0.502,0}{$\downarrow0.090$}  \\
\textit{+ Var-IC}              & 37.55           & 36.57             & 37.35              & \textbf{37.61}                  & \textcolor[rgb]{0,0.502,0}{$\uparrow0.06$} & 0.214          & 0.179            & 0.210             & 0.215                 & \textcolor[rgb]{0,0.502,0}{$\downarrow0.035$}  \\
\textit{+ Var-Prompt}                & 37.04           & 37.14             & 37.25              & 36.77                  & \textcolor[rgb]{0,0.502,0}{$\uparrow0.21$} & 0.229          & 0.139            & 0.191             & 0.219                 & \textcolor[rgb]{0,0.502,0}{$\downarrow0.090$}  \\
\textit{+ Var-Both}                          & 37.04           & 36.67             & 37.15              & 37.50                  & \textcolor[rgb]{0,0.502,0}{$\uparrow0.46$} & 0.229          & \textbf{0.124}            & 0.192             & 0.230                 & \textcolor[rgb]{0,0.502,0}{$\downarrow0.105$}  \\ 
\Xhline{2\arrayrulewidth}

\multirow{3}{*}{\textbf{Systems }}       & \multicolumn{10}{c}{\textbf{Hate Speech }}                                                                                                                                                                                                                       \\
                                         & \multicolumn{5}{c}{\textbf{Macro F1 }}                                                                                       & \multicolumn{5}{c}{\textbf{ECE }}                                                                                                 \\
                                         & \textbf{Ori.} & \textbf{Max} & \textbf{Mean} & \textbf{Majority} & \multicolumn{1}{c}{$\Delta$}                & \textbf{Ori.} & \textbf{Max} & \textbf{Mean} & \textbf{Majority} & $\Delta$                                         \\ 
\hline
\rowcolor[rgb]{0.900,0.900,0.900} ZSL    & 37.08           &                   &                    &                        & \textcolor{red}{$\downarrow0.13$}          & 0.318          &                   &                    &                        & \textcolor[rgb]{0,0.502,0}{$\downarrow0.049$}  \\
\textit{+ Var-Prompt}                & 37.08           & 36.54             & \textbf{36.95}              & \textbf{36.95}                  & \textcolor{red}{$\downarrow0.13$}          & 0.318          & \textbf{0.269}            & 0.302             & 0.320                 & \textcolor[rgb]{0,0.502,0}{$\downarrow0.049$}  \\
\rowcolor[rgb]{0.900,0.900,0.900} ICL    & 40.09           &                   &                    &                        & \textcolor[rgb]{0,0.502,0}{$\uparrow1.10$} & 0.271          &                   &                    &                        & \textcolor[rgb]{0,0.502,0}{$\downarrow0.111$}  \\
\textit{+ Var-IC}              & 40.09           & 40.01             & 39.98              & 40.49                  & \textcolor[rgb]{0,0.502,0}{$\uparrow0.40$} & 0.271          & 0.233            & 0.267             & 0.269                 & \textcolor[rgb]{0,0.502,0}{$\downarrow0.038$}  \\
\textit{+ Var-Prompt}                & 40.09           & 41.03             & \textbf{41.19}              & 41.05                  & \textcolor[rgb]{0,0.502,0}{$\uparrow1.10$} & 0.271          & 0.194            & 0.236             & 0.275                 & \textcolor[rgb]{0,0.502,0}{$\downarrow0.077$}  \\
\textit{+ Var-Both}                          & 40.09           & 39.68             & 40.30              & 40.49                  & \textcolor[rgb]{0,0.502,0}{$\uparrow0.40$} & 0.271          & \textbf{0.160}            & 0.237             & 0.278                 & \textcolor[rgb]{0,0.502,0}{$\downarrow0.111$}  \\
\rowcolor[rgb]{0.900,0.900,0.900} FT     & 58.01           &                   &                    &                        & \textcolor{red}{$\downarrow0.82$}          & 0.354          &                   &                    &                        & \textcolor[rgb]{0,0.502,0}{$\downarrow0.115$}  \\
\textit{+ Var-Prompt}                & 55.92           & 57.16             & 57.17              & \textbf{57.19}                  & \textcolor[rgb]{0,0.502,0}{$\uparrow1.27$} & 0.350          & \textbf{0.239}            & 0.290             & 0.345                 & \textcolor[rgb]{0,0.502,0}{$\downarrow0.111$}  \\
\rowcolor[rgb]{0.900,0.900,0.900} SupICL & 59.48           &                   &                    &                        & \textcolor[rgb]{0,0.502,0}{$\uparrow0.74$} & 0.193          &                   &                    &                        & \textcolor[rgb]{0,0.502,0}{$\downarrow0.078$}  \\
\textit{+ Var-IC}              & 59.48           & 59.98             & 59.82              & 59.83                  & \textcolor[rgb]{0,0.502,0}{$\uparrow0.50$} & 0.193          & 0.141            & 0.179             & 0.191                 & \textcolor[rgb]{0,0.502,0}{$\downarrow0.052$}  \\
\textit{+ Var-Prompt}                & 58.66           & 59.96             & 60.10              & 59.86                  & \textcolor[rgb]{0,0.502,0}{$\uparrow1.44$} & 0.251          & 0.165            & 0.211             & 0.246                 & \textcolor[rgb]{0,0.502,0}{$\downarrow0.086$}  \\
\textit{+ Var-Both}                          & 58.66           & 60.12             & \textbf{60.22}              & 59.97                  & \textcolor[rgb]{0,0.502,0}{$\uparrow1.56$} & 0.251          & \textbf{0.115}            & 0.206             & 0.246                 & \textcolor[rgb]{0,0.502,0}{$\downarrow0.136$}  \\
\Xhline{2\arrayrulewidth}
\end{tabular}
}
\vspace{-1mm}
\caption{Results of self-ensembling with different variations (selection). We mark the cells of baseline systems without self-ensembling and their results in grey. Numbers in \textbf{bold} represents the best values for each learning method. $\Delta$ calculates the difference of performance and calibration error between the original results (Ori.) and the best self-ensembled results, where green denotes better results and red denotes worse results. We refer the readers to Appendix \ref{app:full ensembling results} for full self-ensembling results.}
\label{tab:self ensembling results}
\vspace{-1.5mm}
\end{table*}

%% file: secs/7_ablations.tex
\subsection{Discussions and Ablation Studies}

\rparagraph{Comparison with Classical Calibration Methods}
Compared to Platt and Temperature Scaling \cite{guo2017calibration}, self-ensembling doesn't require any supervised labelled data to tune the parameters for different instruction templates, ICL settings and models. 
In addition, classical calibration methods merely adjust confidence levels while preserving the ranking among choices in classification tasks, making it ineffective in enhancing performance (see Table \ref{apptab:comparison-temp-scaling} in the Appendix). 
As a different perspective, self-ensembling is orthogonal to other calibration methods, allowing other calibration methods to be applied on top of the ensembled results. 

\rparagraph{More ensembling components with \textit{Var-IC} lead to better calibration}
\begin{figure}[!t]
    \centering
    \includegraphics[width=\linewidth]{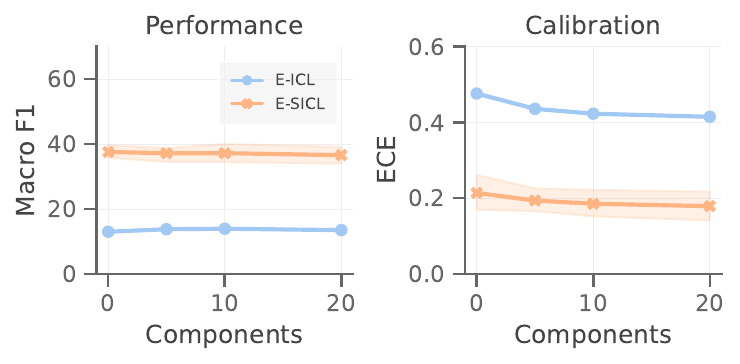}
    \caption{Performance and calibration errors of different number of components with variations in in-context examples on Manifestos. }
    \label{fig:ablation_man_components}
    \vspace{-3mm}
\end{figure}
We explore how varying the number of ensembling components with different in-context examples affects performance and calibration. 
Figure~\ref{fig:ablation_man_components} shows the performance and ECE scores with different components. We observe that although the performances remain comparable, calibration is improved with more components in both ICL and SICL. 
This highlights the effectiveness of the self-ensembling in making the predictions more trustworthy through increased input variations.

\input{tables/appendix/ablation_prompt_num}
\rparagraph{Diversity in \textit{Var-Prompt} influences self-ensembling}
Table~\ref{tab:ablation prompt num} show that
when tuning with more templates, SFT has lower calibration errors whereas those of SICL increase. 
Regarding self-ensembling results, we find that by introducing more prompting templates, self-ensembling yields lower calibration error with SFT, but shows worse calibration with SICL, meanwhile yielding similar performances.
Nonetheless, self-ensembling consistently improves the performance and calibration in each setting. 

\begin{figure}[!t]
    \centering
\vspace{-0mm}
    \includegraphics[width=\linewidth]{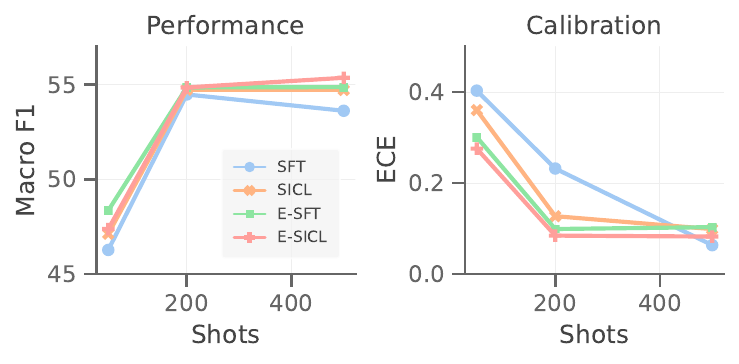}
    \caption{Performance and calibration errors on SST-5 with different numbers of training data. }
    \label{fig:ablation_sst5_shots}
    \vspace{-4mm}
\end{figure}

\rparagraph{The key findings are robust across different training data sizes}
We experiment with larger training data and report the results of SFT and SICL on SST-5 in Figure~\ref{fig:ablation_sst5_shots}. 
We observe that both SFT and SICL yield better performance and lower ECE with more training data, whereas ICL maintains similar performance. 
This indicates that supervised methods can better calibrate the model, and produce more calibrated predictions if provided with sufficient data. 
We also find that self-ensembling remains effective in improving model's performance and mitigating the calibration error on average. 

\rparagraph{The findings hold across different model sizes}
In order to assess the impact of model sizes, we further conduct experiments with Flan-T5\textsubscript{xl}. For full detailed experiment results, we refer the readers to Table \ref{apptab:xl main results} and \ref{apptab:xl ensemble results} in the Appendix. %
We find that a larger model achieves a better calibration score with SFT and SICL than ZSL and ICL. 
We also witness similar behavior of ICL, surpassing SFT and SICL with or without Batch Calibration on seen data, which aligns with previous findings. 
 The results on SST-5 and Hate Speech show that by applying the self-ensembling method, Flan-T5\textsubscript{xl} achieves better performance and lower calibration scores, indicating that the model becomes more `task-specialized'. 
It is also worth noticing that our method is able to improve the performance and decrease the calibration scores on some tasks (Hate Speech) where traditional calibration currently fails. 

%% file: tables/appendix/ablation_prompt_num.tex
\begin{table}[t]
\centering
\resizebox{\linewidth}{!}{%
\begin{tabular}{l|ccccc} 
\Xhline{2\arrayrulewidth}

\multicolumn{6}{c}{\textbf{Hate speech }}                                                                                \\
\multicolumn{1}{l}{\textbf{Metrics}}   & \textbf{Num} & \textbf{SFT} & \textbf{E-SFT} & \textbf{SICL} & \textbf{E-SICL}  \\ 
\hline
\multirow{3}{*}{\textbf{Performance }} & 1            & 58.01        & 58.01          & 59.48         & 59.48            \\
                                       & 4            & 55.92        & 57.16          & 58.66         & 59.96            \\
                                       & 8            & 56.00        & 56.60          & 58.37         & 59.61            \\ 
\hline
\multirow{3}{*}{\textbf{ECE }}         & 1            & 0.354       & 0.354         & 0.193        & 0.193           \\
                                       & 4            & 0.350       & 0.239         & 0.251        & 0.165           \\
                                       & 8            & 0.335       & 0.208         & 0.403        & 0.200           \\ 
\hline
\multicolumn{6}{c}{\textbf{SST-5 }}                                                                                       \\
\multicolumn{1}{l}{\textbf{Metrics}}   & \textbf{Num} & \textbf{SFT} & \textbf{E-SFT} & \textbf{SICL} & \textbf{E-SICL}  \\ 
\hline
\multirow{3}{*}{\textbf{Performance }} & 1            & 46.27        & 46.27          & 47.12         & 47.12            \\
                                       & 4            & 48.11        & 47.43          & 47.99         & 47.70            \\
                                       & 8            & 47.55        & 47.88          & 45.75         & 45.12            \\ 
\hline
\multirow{3}{*}{\textbf{ECE }}         & 1            & 0.403       & 0.403         & 0.360        & 0.360           \\
                                       & 4            & 0.396        & 0.301         & 0.271        & 0.170           \\
                                       & 8            & 0.362        & 0.244         & 0.403        & 0.272           \\
\Xhline{2\arrayrulewidth}
\end{tabular}
}
 \vspace{-1mm}
\caption{Results with different numbers of prompting templates on Hate speech and SST-5.}
\label{tab:ablation prompt num}
\vspace{-2mm}
\end{table}

%% file: secs/conclusion.tex
\vspace{-2mm}
\section{Conclusion}
We have provided a comprehensive analysis of the intricate relationship between in-task performance and calibration across various learning methods in low-resource scenarios. 
Our findings illuminate the nuances of in-task performance and calibration across different task families, meanwhile addressing the inherent miscalibration over all learning methods. 
We have also investigated effective strategies to enhance both aspects simultaneously, offering a viable solution through self-ensembling: it results in more calibrated predictions with comparable or superior task performance. 
We hope that this study will contribute valuable insights into the dynamic landscape of LMs. 
These discoveries also offer practical guidance to practitioners, aiding them in choosing suitable learning paradigms and paving the way for the development of more reliable and high-performing LMs across diverse applications.

%% file: secs/limitations.tex
\section*{Limitations}
Our experimental results are conducted with the Flan-T5 model family, which is an encoder-decoder architecture, where we have not investigated the behaviour of other popular choices of decoder-only models, such as Llama~\cite{llama} in low-resource scenarios, with different learning methods and self-ensembling strategies. Secondly, limited by the training resources, our experiments only consider LMs within the 3B parameter budget. We will endeavor to scale our experiments to cover larger language models as part of future work. Moreover, there is a variety of additional, more sophisticated ensembling methods \cite{MOHAMMED2023757}, where we have only studied the max, mean, and majority vote variants for self-ensembling. In future work, we aim to extend the analysis and the self-ensembling methods to more families of tasks, diverse types of models, and other ensembling techniques.

%% file: appendix/experiment_setup.tex
\section{Experiment Setup}
\label{sec:appendix}
\subsection{Dataset Details}
\label{app:dataset detail}
\rparagraph{SST-2}
The SST-2 dataset, a widely-used benchmark in sentiment analysis, comprises sentences from movie reviews annotated with binary sentiment labels (positive or negative). 
We train the model with the data randomly sampled from the original training set and report the performance on the test set.
We evaluate the model's performance based on accuracy. 

\rparagraph{SST-5}
SST-5, an extension of SST-2, enhances sentiment analysis with five classes: very negative, negative, neutral, positive, and very positive. Derived from movie reviews, this dataset provides a nuanced perspective on sentiment, allowing models to distinguish fine-grained emotional tones.
With all other practices aligned with SST-2, the results are evaluated with micro f1 and macro f1 scores because it has more than 2 labels. 

\rparagraph{RTE}
Recognizing Textual Entailment is a benchmark dataset assessing the task of determining logical entailment between pairs of text snippets. Annotated with binary labels indicating entailment or not, RTE is crucial for evaluating models' logical reasoning abilities. 
We report the accuracy in accordance with other binary classification tasks. 

\rparagraph{ANLI}
Adversarial NLI is a benchmark dataset introducing adversarial examples to challenge models with nuanced reasoning and complex inferences. With labeled sentence pairs denoting entailment, contradiction, or neutrality, ANLI is crucial for assessing models' robustness and generalization in the face of diverse linguistic challenges. 
ANLI has three different rounds of contexts, with later rounds having a better base model, thus being more difficult for the model to distinguish. 
In this paper, we conduct the experiments mainly on the first round, which is easier than other rounds, in order to compare the performance with ICL. 
Since it is a multiclass classification task, we report the performance with micro and macro F1 scores. 
In this paper, we mainly use r1 level data for experiments. 

\rparagraph{NLU++}
NLU++ is a more challenging benchmark in task-oriented dialogue system with more fine-grained domain ontologies and sentences with multiple intents. 
It has two tasks: intent detection and slot labeling, covering the \textit{banking} and \textit{hotels} two domains. 
In this work, we focus on the intent detection task, which is a multi-label classification task and we follow the setting from recent work with state-of-the-art results~\cite{razumovskaia2023sqatin}, which formats it as a binary \texttt{yes/no} classification task. See the cited work for further details.
Regarding the data split, 1,000 sentences from NLU++ were held out for testing and 50 sentences from the leftover 2k+ sentences were sub-sampled for training. 

\rparagraph{Manifestos}
Manifestos was originally created to collect the manifestos of parties from different countries. 
It also includes the analytical variables that indicates the respective categories of the quasi-sentences. 
The corpus have 8 domains overall, which are listed as follows: None (of the below) / Other, External Relations, Freedom and Democracy, Political System, Economy, Welfare and Quality of Life, Fabric of Society, Social Groups. 
In this paper, we use the sentences that only have one golden domain and exclude the ones with multiple labels. 

\rparagraph{Measuring Hate Speech Corpus}
Measuring Hate Speech Corpus, in short Hate speech, contains1 10 constituent ordinal labels and the continuous hate speech score to measure the extent of hate. 
We use the hate speech score as indicator of hate speech in this paper. 
We follow the original division of approximate hate speech provided by the authors, where > 0.5 is approximately hate speech, < -1 is counter or supportive speech, and -1 to 0.5 is neutral or ambiguous.

We only experiment on the intent detection task in the NLU++ bank domain and for ANLI we mainly discuss r1 level data. 
We summarize the training data size, main performance evaluation metrics, and the number of labels for each dataset in Table \ref{tab:dataset-info}. 
We also list the label verbalizers for all datasets in Table \ref{apptab:data verbalizer}.

\input{tables/dataset_info}

\input{tables/appendix/data_verbalizer}

\subsection{Environment Setup}
\label{app:environment}
We mainly use Flan-T5\textsubscript{large} (783M parameters) as the task models for all the datasets. 
We also use Flan-T5\textsubscript{xl} (2.85B parameters) on some of the task to see whether the findings still hold on the larger model. 
For SFT and SICL, we use LoRA~\cite{hu2021lora} to tune Flan-T5\textsubscript{xl}. 
Due to the computational limitations, we can't obtain the results on all the datasets with Flan-T5\textsubscript{xl}. 

All the experiments are conducted on Cambridge High-Performance Clusters with a single A100 (80G) and a 32-core vCPU. 
We release the code and the environment dependencies for reproducible purposes at \url{https://github.com/cambridgeltl/ensembled-sicl}. 

\subsection{Hyperparameters}
\label{app:hyperparameters}
In order to evaluate the model's performance and trustworthiness in low-resource scenarios, we sample a subset of the training set and evaluate it on a fixed set of data as an evaluation and test set. 
For Manifestos, because it has 8 classes and is more expertise in specialized domains (politics, economics and etc.), we use a relatively larger training set to adapt the model to the task itself. 
For Hate Speech, we manually sample the training set and test set ourselves since the corpus didn't provide the split. We randomly sample 1500 data as the fixed test set and 500 examples as the fixed evaluation set. 

All the main experiments are conducted three times with 0, 21, 42 as the random seeds. 
We report the mean values of three runs in the main content. 

Across different learning paradigms (ICL and SICL), we concatenate 3 in-context examples in front of the input for the main experiments. 

For supervised fine-tuning methods, we attach the detailed hyperparameters in Table \ref{tab:hyperparameters} for reproducibility. 
Because tuning the model in the low-resource setting is prone to over-confidence, in order to mitigate the problem, we apply the early stopping with the patience of 5. 

Regarding the configuration hyper-parameters of PEFT, they are listed in Table \ref{tab:peft hyperparameters}.
Unlisted properties use the default values in PEFT implementation from huggingface\footnote{\url{https://huggingface.co/docs/peft/index}}.

\input{tables/appendix/hyperparameters}

\input{tables/appendix/peft_hyperparameters}

%% file: tables/dataset_info.tex
\begin{table}[t]
\renewcommand\arraystretch{1.3}
\resizebox{\columnwidth}{!}{%
\begin{tabular}{cccc}
\Xhline{2\arrayrulewidth}

\textbf{Dataset} & \textbf{Label number} & \textbf{Main Metric} & \textbf{Train size} \\ \hline
SST-2             & 2                     & acc                 & 50                  \\
RTE              & 2                     & acc                 & 50                  \\
ANLI          & 3                     & acc           & 50                  \\
SST5             & 5                     & macro f1           & 50                  \\
NLU++  & 2                     & micro f1      & 50                  \\
Manifestos       & 8                     &  macro f1           & 800                 \\
Hate speech      & 3                     & macro f1           & 50                  \\ \Xhline{2\arrayrulewidth}

\end{tabular}%
}
\caption{Summary of the datasets, main evaluation metric for performance and training data size used for experiment.}
\label{tab:dataset-info}
\end{table}

%% file: tables/appendix/data_verbalizer.tex
\begin{table}
\resizebox{\linewidth}{!}{%
\begin{tabular}{l|l} 
\Xhline{2\arrayrulewidth}
\textbf{Task} & \textbf{Label Verbalizer}                                                                                         \\ 
\hline
SST2          & postive, negative                                                                                                 \\
RTE           & yes, no                                                                                                           \\
ANLI          & yes, maybe, no                                                                                                    \\
SST5          & terrible, bad, neutral, good, great                                                                               \\
NLU++         & yes,no                                                                                                            \\
Manifestos    & \begin{tabular}[c]{@{}l@{}}other, external, democracy, political, \\economy, welfare, fabric, group\end{tabular}  \\
Hate Speech   & support, neutral, hate                                                                                            \\
\Xhline{2\arrayrulewidth}
\end{tabular}
}
\centering
\caption{Label verbalizer for different tasks. }
\label{apptab:data verbalizer}
\end{table}

%% file: tables/appendix/hyperparameters.tex
\begin{table}[]
\resizebox{\columnwidth}{!}{%
\small
\begin{tabular}{lccc}
\Xhline{2\arrayrulewidth}
\textbf{Hyperparameters} & \textbf{ICL}         & \textbf{FT} & \textbf{SupICL} \\ \hline
\multicolumn{4}{c}{\cellcolor[HTML]{EFEFEF}\textbf{SST2}}                       \\ \hline
train batch size         & -                    & 8           & 8               \\
eval batch size          & 64                   & 32          & 32              \\
grad accumulation        & -                    & 1           & 1               \\
learning rate            & -                    & 5e-5        & 5e-5            \\
evaluation per steps     & -                    & 10          & 10              \\
max training epochs      & -                    & 200         & 200             \\
early stopping patience  & -                    & 5           & 5               \\
early stopping metric    & -                    & accuracy    & accuracy        \\ \hline
\multicolumn{4}{c}{\cellcolor[HTML]{EFEFEF}\textbf{RTE}}                        \\ \hline
train batch size         & -                    & 8           & 4               \\
eval batch size          & 64                   & 32          & 32              \\
grad accumulation        & -                    & 1           & 2               \\
learning rate            & -                    & 5e-5        & 5e-5            \\
evaluation per steps     & -                    & 10          & 10              \\
max training epochs      & -                    & 200         & 200             \\
early stopping patience  & -                    & 5           & 5               \\
early stopping metric    & -                    & accuracy    & accuracy        \\ \hline
\multicolumn{4}{c}{\cellcolor[HTML]{EFEFEF}\textbf{ANLI}}                       \\ \hline
train batch size         & -                    & 8           & 4               \\
eval batch size          & 64                   & 32          & 32              \\
grad accumulation        & -                    & 1           & 2               \\
learning rate            & -                    & 5e-5        & 5e-5            \\
evaluation per steps     & -                    & 10          & 10              \\
max training epochs      & -                    & 200         & 200             \\
early stopping patience  & -                    & 5           & 5               \\
early stopping metric    & -                    & accuracy    & accuracy        \\ \hline
\multicolumn{4}{c}{\cellcolor[HTML]{EFEFEF}\textbf{SST5}}                       \\ \hline
train batch size         & -                    & 8           & 8               \\
eval batch size          & 64                   & 32          & 32              \\
grad accumulation        & -                    & 1           & 1               \\
learning rate            & -                    & 5e-5        & 5e-5            \\
evaluation per steps     & -                    & 10          & 10              \\
max training epochs      & -                    & 200         & 200             \\
early stopping patience  & -                    & 5           & 5               \\
early stopping metric    & -                    & macro f1    & macro f1        \\ \hline
\multicolumn{4}{c}{\cellcolor[HTML]{EFEFEF}\textbf{NLU++}}                      \\ \hline
train batch size         & -                    & 16          & 16              \\
eval batch size          & 64                   & 32          & 32              \\
grad accumulation        & -                    & 2           & 2               \\
learning rate            & -                    & 5e-5        & 5e-5            \\
evaluation per steps     & -                    & 500         & 500             \\
max training epochs      & -                    & 200         & 200             \\
early stopping patience  & -                    & 5           & 5               \\
early stopping metric    & -                    & micro f1    & micro f1        \\ \hline
\multicolumn{4}{c}{\cellcolor[HTML]{EFEFEF}\textbf{Manifestos}}                 \\ \hline
train batch size         & -                    & 8           & 4               \\
eval batch size          & 32                   & 32          & 32              \\
grad accumulation        & -                    & 1           & 2               \\
learning rate            & -                    & 5e-5        & 5e-5            \\
evaluation per steps     & -                    & 10          & 10              \\
max training epochs      & -                    & 200         & 200             \\
early stopping patience  & -                    & 5           & 5               \\
early stopping metric    & -                    & macro f1    & macro f1        \\ \hline
\multicolumn{4}{c}{\cellcolor[HTML]{EFEFEF}\textbf{Hate speech}}               \\ \hline
train batch size         & -                    & 8           & 4               \\
eval batch size          & 32                   & 32          & 32              \\
grad accumulation        & -                    & 1           & 2               \\
learning rate            & -                    & 5e-5        & 5e-5            \\
evaluation per steps     & -                    & 10          & 10              \\
max training epochs      & -                    & 200         & 200             \\
early stopping patience  & -                    & 5           & 5               \\
early stopping metric    & -                    & macro f1    & macro f1        \\ \Xhline{2\arrayrulewidth}
\end{tabular}%
}
\caption{Hyper-parameters for each dataset when comparing different learning methods. }
\label{tab:hyperparameters}
\end{table}

%% file: tables/appendix/peft_hyperparameters.tex
\begin{table}[t]
\centering
\resizebox{0.6\columnwidth}{!}{%
\begin{tabular}{lc}
\Xhline{2\arrayrulewidth}
\textbf{Hyperparameters} & \textbf{Values} \\ \hline
r                        & 8               \\
lora alpha               & 32              \\
lora dropout             & 0.05            \\
target modules           & q, v            \\ \Xhline{2\arrayrulewidth}
\end{tabular}%
}
\caption{Hyper-parameters for PEFT with FlanT5-xl. }
\label{tab:peft hyperparameters}
\end{table}

%% file: appendix/full_results.tex
\section{Full Experiment Results}
\label{sec:full results}

To solidify the empirical findings, in this section, we present full experiment results with more metrics in addition to the table in the main content for readers' reference. 

\subsection{Results of different learning methods}
\label{app:full main results}
Table \ref{apptab:main full results} shows the full results on all 7 datasets. 
We report the accuracy, micro f1, and macro f1 as the performance metrics.
We report ECE as the measurement of calibration. 
We also include NLL and IE as supplementary uncertainty metrics. 
We find that on SST-2 and RTE, the model achieves comparable or even better performance with ZSL and ICL than SFT and SICL. 
In the meantime, the predictions have a relatively high ECE, indicating that on these two datasets, the model has the issue of miscalibration. 
On ANLI, Manifestos, Hate speech, and NLU++, with SFT and SICL the model has lower calibration error than ICL and achieves better performance in both performance metrics.

In addition to the original results, we include the Batch Calibration results across all the datasets. 
On SST-5 and ANLI, although ZSL and ICL achieve similar micro f1 scores, there is still a gap in the original macro f1 scores between ZSL/ICL and SFT/SICL. 
However, after applying Batch Calibration, we find that ZSL/ICL has a comparable macro f1 score to SFT/SICL. 
On Manifestos, Hate speech, and NLU++, we don't observe comparable performance between ZSL/ICL and SFT/SICL either with or without Batch Calibration.

\input{tables/appendix/main_full_results}

\subsection{Results of self-ensembling}
\label{app:full ensembling results}
Table \ref{apptab:self ensembling full results} shows the self-ensembling results across 4 datasets. 
We exclude the seen datasets (SST-2 and RTE) for fair evaluation, as well as NLU++ since it's almost well-calibrated with supervised tuning. 
We still include ANLI for comparison even though it is included during pre-training.
We report the mean values of the results with 3 different random seeds (0, 21, 42). 

From the perspective of performance, we find that on SST-5, Manifestos, and Hate speech, self-ensembled results achieve slightly better performances on average and show positive improvements with each learning method. 
On ANLI, we observe no significant improvement in the accuracy of self-ensembled results and the decreases in performance are trivial as well. 
However, from the perspective of calibration, we find that self-ensembling with max probability consistently decreases the calibration over all settings, as shown in Figure \ref{fig:eceplots}. 
Introducing variations in both in-context examples and prompting templates yields the lowest calibration error in all experiments. 

Among different ensembling methods, we find that majority vote can achieve better performances sometimes but it doesn't help to reduce the calibration error or even make it worse. 
Mean probability and max probability are able to improve the performance meanwhile reducing the calibration error. 
The empirical experiment results suggest that although majority vote as a widely used ensemble method achieves better performance, it is worth noting that it may deliver unfaithful predictions, which is not preferred in real application.

\input{tables/appendix/ensemble_full_results}
\input{figs/latex/ece_plots}

%% file: tables/appendix/main_full_results.tex
\begin{table*}
\centering
\renewcommand\arraystretch{1.2}
\resizebox{\linewidth}{!}{%
\begin{tabular}{ll|cccc|cccc} 
\Xhline{2\arrayrulewidth}
\multicolumn{2}{l|}{\multirow{2}{*}{\textbf{Evaluation Metrics }}}                                                                 & \multicolumn{4}{c|}{\textbf{SST2 }}                                                                       & \multicolumn{4}{c}{\textbf{RTE }}                                                                          \\
\multicolumn{2}{l|}{}                                                                                                              & \textbf{ZSL} & \textbf{ICL}                 & \textbf{SFT}                  & \textbf{SICL}              & \textbf{ZSL} & \textbf{ICL}                 & \textbf{SFT}                  & \textbf{SICL}               \\ 
\hline
\multirow{2}{*}{\textbf{Performance }}                                                                         & \textbf{acc}      & 94.67       & 95.22\textsubscript{0.12}           & 95.61\textsubscript{0.20}           & 95.63\textsubscript{0.29}           & 86.64        & 88.45\textsubscript{0}       & 88.81\textsubscript{0.29}    & 88.57\textsubscript{0.45}     \\
                                                                                                               & \textbf{macro f1} & -            & -                            & -                            & -                            & -            & -                            & -                            & -                             \\
\rowcolor[rgb]{0.900,0.900,0.900} {\cellcolor[rgb]{0.900,0.900,0.900}}                                         & \textbf{acc}      & 95.50       & 95.95\textsubscript{0.19}           & 95.61\textsubscript{0.20}           & 95.63\textsubscript{0.29}           & 89.53        & 90.25\textsubscript{0.51}    & 89.17\textsubscript{0.29}    & 87.97\textsubscript{0.45}     \\
\rowcolor[rgb]{0.900,0.900,0.900} \multirow{-2}{*}{{\cellcolor[rgb]{0.900,0.900,0.900}}\textbf{\textit{+ calibrated} }} & \textbf{macro f1} & -            & -                            & -                            & -                            & -            & -                            & -                            & -                             \\
\multirow{3}{*}{\textbf{Trustworthiness }}                                                                     & \textbf{ECE}      & 0.9069       & 0.9149\textsubscript{0.0014}           & 0.9408\textsubscript{0.0113}           & 0.9449\textsubscript{0.0129}           & 0.8092       & 0.8150\textsubscript{0.0030} & 0.8418\textsubscript{0.0224} & 0.8759\textsubscript{0.0025}  \\
                                                                                                               & \textbf{NLL}      & 0.1465       & 0.1347\textsubscript{0.0006}           & 0.2146\textsubscript{0.0962}           & 0.2710\textsubscript{0.1396}           & 0.3438       & 0.2910\textsubscript{0.0037} & 0.4421\textsubscript{0.2133} & 1.1234\textsubscript{0.1133}  \\
                                                                                                               & \textbf{IE}       & 0.0615       & 0.0599\textsubscript{0.0003}           & 0.0215\textsubscript{0.0149}           & 0.0154\textsubscript{0.0124}           & 0.0977       & 0.0975\textsubscript{0.0002} & 0.0630\textsubscript{0.0304} & 0.0153\textsubscript{0.0030}  \\
\rowcolor[rgb]{0.900,0.900,0.900} \textbf{\textit{+ calibrated}}                                                        & \textbf{ECE}      & 0.7877       & 0.7961\textsubscript{0.0016} & 0.8040\textsubscript{0.0056} & 0.8064\textsubscript{0.0096} & 0.6831       & 0.6991\textsubscript{0.0054} & 0.7022\textsubscript{0.0061} & 0.7111\textsubscript{0.0045}  \\ 
\hline
\multicolumn{2}{l|}{\multirow{2}{*}{\textbf{Evaluation Metrics }}}                                                                 & \multicolumn{4}{c|}{\textbf{SST5 }}                                                                       & \multicolumn{4}{c}{\textbf{ANLI }}                                                                         \\
\multicolumn{2}{l|}{}                                                                                                              & \textbf{ZSL} & \textbf{ICL}                 & \textbf{SFT}                  & \textbf{SICL}              & \textbf{ZSL} & \textbf{ICL}                 & \textbf{SFT}                  & \textbf{SICL}               \\ 
\hline
\multirow{2}{*}{\textbf{Performance }}                                                                         & \textbf{micro f1} & 52.58       & 50.48\textsubscript{0.15}           & 50.59\textsubscript{1.38}           & 54.25\textsubscript{0.46}           & 52.30        & 52.17\textsubscript{0.47}    & 61.63\textsubscript{1.68}    & 63.90\textsubscript{0.14}     \\
                                                                                                               & \textbf{macro f1} & 42.00       & 37.59\textsubscript{0.23}           & 46.27\textsubscript{2.09}           & 47.12\textsubscript{0.0193}           & 42.07        & 42.06\textsubscript{0.39}    & 61.17\textsubscript{1.76}    & 63.53\textsubscript{0.43}     \\
\rowcolor[rgb]{0.900,0.900,0.900} {\cellcolor[rgb]{0.900,0.900,0.900}}                                         & \textbf{micro f1} & 50.05       & 50.89\textsubscript{0.86}           & 50.35\textsubscript{0.50}           & 49.91\textsubscript{0.21}           & 62.30        & 61.27\textsubscript{0.82}    & 62.47\textsubscript{1.60}    & 64.50\textsubscript{0.41}     \\
\rowcolor[rgb]{0.900,0.900,0.900} \multirow{-2}{*}{{\cellcolor[rgb]{0.900,0.900,0.900}}\textbf{\textit{+ calibrated} }} & \textbf{macro f1} & 48.98       & 49.80\textsubscript{0.91}           & 50.43\textsubscript{0.68}           & 49.89\textsubscript{0.26}           & 61.98        & 61.15\textsubscript{0.83}    & 62.36\textsubscript{1.67}    & 64.46\textsubscript{0.40}     \\
\multirow{3}{*}{\textbf{Trustworthiness }}                                                                     & \textbf{ECE}      & 0.1416       & 0.1833\textsubscript{0.0020}           & 0.4030\textsubscript{0.0321}           & 0.3602\textsubscript{0.0250}           & 0.3555       & 0.3511\textsubscript{0.0050} & 0.3161\textsubscript{0.0230} & 0.2803\textsubscript{0.0108}  \\
                                                                                                               & \textbf{NLL}      & 1.1762       & 1.2261\textsubscript{0.0021}           & 3.1986\textsubscript{1.4010}           & 2.3390\textsubscript{0.2827}           & 4.7484       & 3.8961\textsubscript{0.0233} & 2.2599\textsubscript{0.3255} & 1.9226\textsubscript{0.2912}  \\
                                                                                                               & \textbf{IE}       & 0.1599       & 0.1522\textsubscript{0.0001}           & 0.0454\textsubscript{0.0209}           & 0.0489\textsubscript{0.0107}           & 0.0945       & 0.0987\textsubscript{0.0001} & 0.0587\textsubscript{0.0070} & 0.0663\textsubscript{0.0113}  \\
\rowcolor[rgb]{0.900,0.900,0.900} \textbf{\textit{+ calibrated}}                                                        & \textbf{ECE}      & 0.1010       & 0.1027\textsubscript{0.0084} & 0.0720\textsubscript{0.0129} & 0.0456\textsubscript{0.0035} & 0.0709       & 0.0430\textsubscript{0.0079} & 0.0626\textsubscript{0.0145} & 0.0495\textsubscript{0.0085}  \\ 
\hline
\multicolumn{2}{l|}{\multirow{2}{*}{\textbf{Evaluation Metrics }}}                                                                 & \multicolumn{4}{c|}{\textbf{Manifestos }}                                                                 & \multicolumn{4}{c}{\textbf{Hate speech }}                                                                  \\
\multicolumn{2}{l|}{}                                                                                                              & \textbf{ZSL} & \textbf{ICL}                 & \textbf{SFT}                  & \textbf{SICL}              & \textbf{ZSL} & \textbf{ICL}                 & \textbf{SFT}                  & \textbf{SICL}               \\ 
\hline
\multirow{2}{*}{\textbf{Performance }}                                                                         & \textbf{micro f1} & 20.87       & 19.29\textsubscript{0.16}           & 37.54\textsubscript{1.10}           & 38.12\textsubscript{2.01}           & 39.67       & 40.18\textsubscript{0.14}           & 59.67\textsubscript{0.47}           & 61.33\textsubscript{2.05}            \\
                                                                                                               & \textbf{macro f1} & 14.50       & 13.01\textsubscript{0.19}           & 35.76\textsubscript{1.23}           & 37.55\textsubscript{1.61}           & 37.08       & 40.09\textsubscript{0.08}           & 58.01\textsubscript{1.01}           & 59.48\textsubscript{1.79}            \\
\rowcolor[rgb]{0.900,0.900,0.900} {\cellcolor[rgb]{0.900,0.900,0.900}}                                         & \textbf{micro f1} & 33.63       & 31.00\textsubscript{0.37}           & 38.58\textsubscript{0.46}           & 38.33\textsubscript{0.33}           & 43.87       & 45.11\textsubscript{0.46}           & 59.89\textsubscript{0.17}           & 59.58\textsubscript{3.26}            \\
\rowcolor[rgb]{0.900,0.900,0.900} \multirow{-2}{*}{{\cellcolor[rgb]{0.900,0.900,0.900}}\textbf{\textit{+ calibrated} }} & \textbf{macro f1} & 30.86       & 29.15\textsubscript{0.53}           & 37.57\textsubscript{1.05}           & 37.83\textsubscript{0.41}           & 40.86       & 42.36\textsubscript{0.42}           & 58.06\textsubscript{0.27}           & 57.81\textsubscript{3.03}            \\
\multirow{3}{*}{\textbf{Trustworthiness }}                                                                     & \textbf{ECE}      & 0.4319       & 0.4760\textsubscript{0.0018}           & 0.2005\textsubscript{0.0723}           & 0.2138\textsubscript{0.0376}           & 0.3175       & 0.2708\textsubscript{0.0024}           & 0.3541\textsubscript{0.0364}           & 0.1928\textsubscript{0.1131}            \\
                                                                                                               & \textbf{NLL}      & 3.7344       & 3.9423\textsubscript{0.0091}           & 2.0725\textsubscript{0.1637}           & 2.0264\textsubscript{0.0751}           & 1.3836       & 1.2136\textsubscript{0.0057}           & 4.4720\textsubscript{2.4332}           & 2.0827\textsubscript{1.6186}            \\
                                                                                                               & \textbf{IE}       & 0.1141       & 0.1006\textsubscript{0.0001}           & 0.1469\textsubscript{0.0182}           & 0.1446\textsubscript{0.0104}           & 1.0112       & 0.2426\textsubscript{0.0007}           & 0.0391\textsubscript{0.0286}           & 0.1447\textsubscript{0.0963}            \\
\rowcolor[rgb]{0.900,0.900,0.900} \textbf{\textit{+ calibrated}}                                                        & \textbf{ECE}      & 0.0356       & 0.0721\textsubscript{0.0020} & 0.0487\textsubscript{0.0120} & 0.0654\textsubscript{0.0043} & 0.1107       & 0.1173\textsubscript{0.0045} & 0.0646\textsubscript{0.0274} & 0.0453\textsubscript{0.0114}  \\ 
\hline
\multicolumn{2}{l|}{\multirow{2}{*}{\textbf{Evaluation Metrics }}}                                                                 & \multicolumn{4}{c|}{\textbf{NLU++ }}                                                                      &              &                              &                              &                               \\
\multicolumn{2}{l|}{}                                                                                                              & \textbf{ZSL} & \textbf{ICL}                 & \textbf{SFT}                  & \textbf{SICL}              &              &                              &                              &                               \\ 
\hline
\multirow{2}{*}{\textbf{Performance }}                                                                         & \textbf{micro f1} & 29.2         & 40.11\textsubscript{0.09}    & 79.98\textsubscript{0.59}    & 80.76\textsubscript{0.31}    &              &                              &                              &                               \\
                                                                                                               & \textbf{macro f1} & 40.26      & 51.96\textsubscript{0.04}    & 80.58\textsubscript{1.00}    & 80.49\textsubscript{0.12}    &              &                              &                              &                               \\
\rowcolor[rgb]{0.900,0.900,0.900} {\cellcolor[rgb]{0.900,0.900,0.900}}                                         & \textbf{micro f1} & 11.18      & 12.25\textsubscript{0.02}    & 16.45\textsubscript{0.72}    & 21.00\textsubscript{4.78}    &              &                              &                              &                               \\
\rowcolor[rgb]{0.900,0.900,0.900} \multirow{-2}{*}{{\cellcolor[rgb]{0.900,0.900,0.900}}\textbf{\textit{+ calibrated} }} & \textbf{macro f1} & 11.35      & 12.56\textsubscript{0.05}    & 21.27\textsubscript{2.05}    & 27.07\textsubscript{3.58}    &              &                              &                              &                               \\
\multirow{3}{*}{\textbf{Trustworthiness }}                                                                     & \textbf{ECE}      & 0.2311       & 0.1291\textsubscript{0.0001} & 0.0112\textsubscript{0.0007} & 0.0020\textsubscript{0.0007} &              &                              &                              &                               \\
                                                                                                               & \textbf{NLL}      & 0.3435       & 0.2140\textsubscript{0.0001} & 0.1305\textsubscript{0.0348} & 0.0839\textsubscript{0.0127} &              &                              &                              &                               \\
                                                                                                               & \textbf{IE}       & 0.2268       & 0.1419\textsubscript{0.0001} & 0.0014\textsubscript{0.0007} & 0.0020\textsubscript{0.0007} &              &                              &                              &                               \\
\rowcolor[rgb]{0.900,0.900,0.900} \textbf{\textit{+ calibrated}}                                                        & \textbf{ECE}      & 0.4679       & 0.4559\textsubscript{0.0001} & 0.4183\textsubscript{0.0038} & 0.4098\textsubscript{0.0049} &              &                              &                              &                               \\
\Xhline{2\arrayrulewidth}
\end{tabular}
}
\caption{Full experiment results across 7 datasets with different learning methods. We report the mean value for 3 runs with different random seeds and list the variance in the subscripts. We color the Batch Calibration results in grey. }
\label{apptab:main full results}
\end{table*}

%% file: tables/appendix/ensemble_full_results.tex
\begin{table*}
\centering
\renewcommand\arraystretch{1.0}
\resizebox{0.9\linewidth}{!}{%
\begin{tabular}{l|ccccc|ccccc} 
\Xhline{2\arrayrulewidth}
\multirow{3}{*}{\textbf{Systems }}       & \multicolumn{10}{c}{\textbf{SST5 }}                                                                                                                                                                                                                              \\
                                         & \multicolumn{5}{c}{\textbf{Macro F1 }}                                                                                       & \multicolumn{5}{c}{\textbf{ECE }}                                                                                                 \\
                                         & \textbf{Ori.} & \textbf{Max} & \textbf{Mean} & \textbf{Majority} & \multicolumn{1}{c}{$\Delta$}                & \textbf{Ori.} & \textbf{Max} & \textbf{Mean} & \textbf{Majority} & $\Delta$                                         \\ 
\hline
\rowcolor[rgb]{0.900,0.900,0.900} ZSL    & 42.00           &                   &                    &                        & \textcolor[rgb]{0,0.502,0}{$\uparrow4.79$} & 0.1416          &                   &                    &                        & \textcolor[rgb]{0,0.502,0}{$\downarrow0.0634$}  \\
\textit{+ Var-Prompt}                & 42.00           & \textbf{46.79}             & 45.70              & 44.82                  & \textcolor[rgb]{0,0.502,0}{$\uparrow4.79$} & 0.1416          & \textbf{0.0782}            & 0.1132             & 0.1374                 & \textcolor[rgb]{0,0.502,0}{$\downarrow0.0634$}  \\
\rowcolor[rgb]{0.900,0.900,0.900} ICL    & 37.59           &                   &                    &                        & \textcolor[rgb]{0,0.502,0}{$\uparrow0.16$} & 0.1833          &                   &                    &                        & \textcolor[rgb]{0,0.502,0}{$\downarrow0.0911$}  \\
\textit{+ Var-IC}              & 37.59           & \textbf{37.75}             & 37.33              & 37.26                  & \textcolor[rgb]{0,0.502,0}{$\uparrow0.16$} & 0.1833          & 0.1198            & 0.1749             & 0.1857                 & \textcolor[rgb]{0,0.502,0}{$\downarrow0.0635$}  \\
\textit{+ Var-Prompt}                & 37.59           & 37.13             & 36.52              & 36.83                  & \textcolor{red}{$\downarrow0.46$}          & 0.1833          & 0.1363            & 0.1838             & 0.2025                 & \textcolor[rgb]{0,0.502,0}{$\downarrow0.0470$}  \\
\textit{+ Var-Both}                          & 37.59           & 33.82             & 35.33              & 35.78                  & \textcolor{red}{$\downarrow1.81$}          & 0.1833          & \textbf{0.0955}            & 0.1832             & 0.2107                 & \textcolor[rgb]{0,0.502,0}{$\downarrow0.0911$}  \\
\rowcolor[rgb]{0.900,0.900,0.900} FT     & 46.27           &                   &                    &                        & \textcolor[rgb]{0,0.502,0}{$\uparrow2.08$} & 0.4030          &                   &                    &                        & \textcolor[rgb]{0,0.502,0}{$\downarrow0.1022$}  \\
\textit{+ Var-Prompt}                & 48.11           & 47.43             & \textbf{48.35}              & 48.33                  & \textcolor[rgb]{0,0.502,0}{$\uparrow0.24$} & 0.3960          & \textbf{0.3008}            & 0.3466             & 0.3973                 & \textcolor[rgb]{0,0.502,0}{$\downarrow0.0952$}  \\
\rowcolor[rgb]{0.900,0.900,0.900} SupICL & 47.12           &                   &                    &                        & \textcolor[rgb]{0,0.502,0}{$\uparrow0.79$} & 0.3602          &                   &                    &                        & \textcolor[rgb]{0,0.502,0}{$\downarrow0.2402$}  \\
\textit{+ Var-IC}              & 47.12           & 47.30           & 47.37              & 47.31                  & \textcolor[rgb]{0,0.502,0}{$\uparrow0.25$} & 0.3602          & 0.2755            & 0.3428             & 0.3615                 & \textcolor[rgb]{0,0.502,0}{$\downarrow0.0847$}  \\
\textit{+ Var-Prompt}                & 47.99           & 47.70           & 47.88              & \textbf{47.91}                  & \textcolor{red}{$\downarrow0.08$}          & 0.2714          & 0.1698            & 0.2342             & 0.2801                 & \textcolor[rgb]{0,0.502,0}{$\downarrow0.1016$}  \\
\textit{+ Var-Both}                          & 47.99           & 47.18             & 47.56              & 47.54                  & \textcolor{red}{$\downarrow0.43$}          & 0.2714          & \textbf{0.1200}            & 0.2296             & 0.2791                 & \textcolor[rgb]{0,0.502,0}{$\downarrow0.1514$}  \\ 
\hline
\multirow{3}{*}{\textbf{Systems }}       & \multicolumn{10}{c}{\textbf{Manifestos }}                                                                                                                                                                                                                        \\
                                         & \multicolumn{5}{c}{\textbf{Macro F1 }}                                                                                       & \multicolumn{5}{c}{\textbf{ECE }}                                                                                                 \\
                                         & \textbf{Ori.} & \textbf{Max} & \textbf{Mean} & \textbf{Majority} & \multicolumn{1}{c}{$\Delta$}                & \textbf{Ori.} & \textbf{Max} & \textbf{Mean} & \textbf{Majority} & $\Delta$                                         \\ 
\hline
\rowcolor[rgb]{0.900,0.900,0.900} ZSL    & 14.50           &                   &                    &                        & \textcolor[rgb]{0,0.502,0}{$\uparrow0.79$} & 0.4319          &                   &                    &                        & \textcolor[rgb]{0,0.502,0}{$\downarrow0.1699$}  \\
\textit{+ Var-Prompt}                & 14.50           & 13.69             & 12.93              & \textbf{15.29}                  & \textcolor[rgb]{0,0.502,0}{$\uparrow0.79$} & 0.4319          & \textbf{0.2620}            & 0.3349             & 0.4374                 & \textcolor[rgb]{0,0.502,0}{$\downarrow0.1699$}  \\
\rowcolor[rgb]{0.900,0.900,0.900} ICL    & 13.01           &                   &                    &                        & \textcolor[rgb]{0,0.502,0}{$\uparrow0.68$} & 0.4760          &                   &                    &                        & \textcolor[rgb]{0,0.502,0}{$\downarrow0.2828$}  \\
\textit{+ Var-IC}              & 13.01           & 13.50             & 13.46              & \textbf{13.69}                  & \textcolor[rgb]{0,0.502,0}{$\uparrow0.68$} & 0.4760          & 0.4146            & 0.4645             & 0.4689                 & \textcolor[rgb]{0,0.502,0}{$\downarrow0.0614$}  \\
\textit{+ Var-Prompt}                & 13.01           & 13.25             & 11.67              & 11.53                  & \textcolor[rgb]{0,0.502,0}{$\uparrow0.24$} & 0.4760          & 0.2682            & 0.3661             & 0.4718                 & \textcolor[rgb]{0,0.502,0}{$\downarrow0.2078$}  \\
\textit{+ Var-Both}                          & 13.01           & 11.19             & 11.50              & 11.21                  & \textcolor{red}{$\downarrow1.51$}          & 0.4760          & \textbf{0.1932}            & 0.3537             & 0.4796                 & \textcolor[rgb]{0,0.502,0}{$\downarrow0.2828$}  \\
\rowcolor[rgb]{0.900,0.900,0.900} FT     & 35.76           &                   &                    &                        & \textcolor[rgb]{0,0.502,0}{$\uparrow0.73$} & 0.2005          &                   &                    &                        & \textcolor[rgb]{0,0.502,0}{$\downarrow0.1343$}  \\
\textit{+ Var-Prompt}                & 35.39           & \textbf{36.49}             & 35.66              & 34.91                  & \textcolor[rgb]{0,0.502,0}{$\uparrow1.10$} & 0.1435          & \textbf{0.0662}            & 0.1049             & 0.1352                 & \textcolor[rgb]{0,0.502,0}{$\downarrow0.0773$}  \\
\rowcolor[rgb]{0.900,0.900,0.900} SupICL & 37.55           &                   &                    &                        & \textcolor[rgb]{0,0.502,0}{$\uparrow0.06$} & 0.2138          &                   &                    &                        & \textcolor[rgb]{0,0.502,0}{$\downarrow0.0902$}  \\
\textit{+ Var-IC}              & 37.55           & 36.57             & 37.35              & \textbf{37.61}                  & \textcolor[rgb]{0,0.502,0}{$\uparrow0.06$} & 0.2138          & 0.1789            & 0.2096             & 0.2152                 & \textcolor[rgb]{0,0.502,0}{$\downarrow0.0349$}  \\
\textit{+ Var-Prompt}                & 37.04           & 37.14             & 37.25              & 36.77                  & \textcolor[rgb]{0,0.502,0}{$\uparrow0.21$} & 0.2285          & 0.1388            & 0.1912             & 0.2190                 & \textcolor[rgb]{0,0.502,0}{$\downarrow0.0897$}  \\
\textit{+ Var-Both}                          & 37.04           & 36.67             & 37.15              & 37.50                  & \textcolor[rgb]{0,0.502,0}{$\uparrow0.46$} & 0.2285          & \textbf{0.1236}            & 0.1918             & 0.2303                 & \textcolor[rgb]{0,0.502,0}{$\downarrow0.1049$}  \\ 
\hline
\multirow{3}{*}{\textbf{Systems }}       & \multicolumn{10}{c}{\textbf{Hate Speech }}                                                                                                                                                                                                                       \\
                                         & \multicolumn{5}{c}{\textbf{Macro F1 }}                                                                                       & \multicolumn{5}{c}{\textbf{ECE }}                                                                                                 \\
                                         & \textbf{Ori.} & \textbf{Max} & \textbf{Mean} & \textbf{Majority} & \multicolumn{1}{c}{$\Delta$}                & \textbf{Ori.} & \textbf{Max} & \textbf{Mean} & \textbf{Majority} & $\Delta$                                         \\ 
\hline
\rowcolor[rgb]{0.900,0.900,0.900} ZSL    & 37.08           &                   &                    &                        & \textcolor{red}{$\downarrow0.13$}          & 0.3175          &                   &                    &                        & \textcolor[rgb]{0,0.502,0}{$\downarrow0.0489$}  \\
\textit{+ Var-Prompt}                & 37.08           & 36.54             & \textbf{36.95}              & \textbf{36.95}                  & \textcolor{red}{$\downarrow0.13$}          & 0.3175          & \textbf{0.2686}            & 0.3024             & 0.3200                 & \textcolor[rgb]{0,0.502,0}{$\downarrow0.0489$}  \\
\rowcolor[rgb]{0.900,0.900,0.900} ICL    & 40.09           &                   &                    &                        & \textcolor[rgb]{0,0.502,0}{$\uparrow1.10$} & 0.2708          &                   &                    &                        & \textcolor[rgb]{0,0.502,0}{$\downarrow0.1112$}  \\
\textit{+ Var-IC}              & 40.09           & 40.01             & 39.98              & 40.49                  & \textcolor[rgb]{0,0.502,0}{$\uparrow0.40$} & 0.2708          & 0.2332            & 0.2668             & 0.2694                 & \textcolor[rgb]{0,0.502,0}{$\downarrow0.0376$}  \\
\textit{+ Var-Prompt}                & 40.09           & 41.03             & \textbf{41.19}              & 41.05                  & \textcolor[rgb]{0,0.502,0}{$\uparrow1.10$} & 0.2708          & 0.1944            & 0.2359             & 0.2745                 & \textcolor[rgb]{0,0.502,0}{$\downarrow0.0764$}  \\
\textit{+ Var-Both}                          & 40.09           & 39.68             & 40.30              & 40.49                  & \textcolor[rgb]{0,0.502,0}{$\uparrow0.40$} & 0.2708          & \textbf{0.1596}            & 0.2366             & 0.2776                 & \textcolor[rgb]{0,0.502,0}{$\downarrow0.1112$}  \\
\rowcolor[rgb]{0.900,0.900,0.900} FT     & 58.01           &                   &                    &                        & \textcolor{red}{$\downarrow0.82$}          & 0.3541          &                   &                    &                        & \textcolor[rgb]{0,0.502,0}{$\downarrow0.1155$}  \\
\textit{+ Var-Prompt}                & 55.92           & 57.16             & 57.17              & \textbf{57.19}                  & \textcolor[rgb]{0,0.502,0}{$\uparrow1.27$} & 0.3502          & \textbf{0.2386}            & 0.2899             & 0.3448                 & \textcolor[rgb]{0,0.502,0}{$\downarrow0.1116$}  \\
\rowcolor[rgb]{0.900,0.900,0.900} SupICL & 59.48           &                   &                    &                        & \textcolor[rgb]{0,0.502,0}{$\uparrow0.74$} & 0.1928          &                   &                    &                        & \textcolor[rgb]{0,0.502,0}{$\downarrow0.0777$}  \\
\textit{+ Var-IC}              & 59.48           & 59.98             & 59.82              & 59.83                  & \textcolor[rgb]{0,0.502,0}{$\uparrow0.50$} & 0.1928          & 0.1413            & 0.1789             & 0.1905                 & \textcolor[rgb]{0,0.502,0}{$\downarrow0.0515$}  \\
\textit{+ Var-Prompt}                & 58.66           & 59.96             & 60.10              & 59.86                  & \textcolor[rgb]{0,0.502,0}{$\uparrow1.44$} & 0.2507          & 0.1648            & 0.2111             & 0.2457                 & \textcolor[rgb]{0,0.502,0}{$\downarrow0.0859$}  \\
\textit{+ Var-Both}                          & 58.66           & 60.12             & \textbf{60.22}              & 59.97                  & \textcolor[rgb]{0,0.502,0}{$\uparrow1.56$} & 0.2507          & \textbf{0.1151}            & 0.2057             & 0.2455                 & \textcolor[rgb]{0,0.502,0}{$\downarrow0.1356$}  \\
\hline
\multirow{3}{*}{\textbf{Systems }}       & \multicolumn{10}{c}{\textbf{ANLI }}                                                                                                                                                                                                                                      \\
                                         & \multicolumn{5}{c}{\textbf{Acc }}                                                                                                & \multicolumn{5}{c}{\textbf{ECE }}                                                                                                     \\
                                         & \textbf{Ori.} & \textbf{Max} & \textbf{Mean} & \textbf{Majority} & \multicolumn{1}{c}{\textbf{$\Delta$}}      & \textbf{Ori.} & \textbf{Max} & \textbf{Mean} & \textbf{Majority} & \textbf{$\Delta$}                               \\ 
\hline
\rowcolor[rgb]{0.900,0.900,0.900} ZSL    & 52.30           &                   &                    &                        & \textcolor{red}{$\downarrow0.60$}          & 0.3555          &                   &                    &                        & \textcolor[rgb]{0,0.502,0}{$\downarrow0.0133$}  \\
\textit{+ Var-Prompt}                & 52.30           & 51.70             & 51.60              & 51.60                  & \textcolor{red}{$\downarrow0.60$}          & 0.3555          & \textbf{0.3422}            & 0.3603             & 0.3640                 & \textcolor[rgb]{0,0.502,0}{$\downarrow0.0133$}  \\
\rowcolor[rgb]{0.900,0.900,0.900} ICL    & 52.17           &                   &                    &                        & \textcolor{red}{$\downarrow0.01$}          & 0.3511          &                   &                    &                        & \textcolor[rgb]{0,0.502,0}{$\downarrow0.0204$}  \\
\textit{+ Var-IC}              & 52.17           & 52.10             & 52.10              & 52.10                  & \textcolor{red}{$\downarrow0.07$}          & 0.3511          & 0.3347            & 0.3506             & 0.3521                 & \textcolor[rgb]{0,0.502,0}{$\downarrow0.0164$}  \\
\textit{+ Var-Prompt}                & 52.17           & 52.03             & 52.07              & \textbf{52.16}                  & \textcolor{red}{$\downarrow0.01$}          & 0.3511          & 0.3375            & 0.3490              & 0.3506                 & \textcolor[rgb]{0,0.502,0}{$\downarrow0.0136$}  \\
\textit{+ Var-Both}                          & 52.17           & \textbf{51.70}             & 51.80              & 51.80                  & \textcolor{red}{$\downarrow0.37$}          & 0.3511          & \textbf{0.3307}            & 0.3510              & 0.3544                 & \textcolor[rgb]{0,0.502,0}{$\downarrow0.0204$}  \\
\rowcolor[rgb]{0.900,0.900,0.900} FT     & 61.63           &                   &                    &                        & \textcolor{red}{$\downarrow0.07$}          & 0.3161          &                   &                    &                        & \textcolor[rgb]{0,0.502,0}{$\downarrow0.0271$}  \\
\textit{+ Var-Prompt}                & 61.53           & \textbf{61.56}             & 61.37              & 61.37                  & \textcolor[rgb]{0,0.502,0}{$\uparrow0.03$} & 0.3160          & \textbf{0.2890}            & 0.3101             & 0.3184                 & \textcolor[rgb]{0,0.502,0}{$\downarrow0.0270$}  \\
\rowcolor[rgb]{0.900,0.900,0.900} SupICL & 63.90           &                   &                    &                        & \textcolor[rgb]{0,0.502,0}{$\uparrow0.47$} & 0.2803          &                   &                    &                        & \textcolor[rgb]{0,0.502,0}{$\downarrow0.0303$}  \\
\textit{+ Var-IC}              & 63.90           & 63.87             & 63.73              & 63.83                  & \textcolor{red}{$\downarrow0.03$}          & 0.2803          & 0.2293            & 0.2740             & 0.2839                 & \textcolor[rgb]{0,0.502,0}{$\downarrow0.0510$}  \\
\textit{+ Var-Prompt}                & 63.97           & 64.33             & \textbf{64.37}              & \textbf{64.37}                  & \textcolor[rgb]{0,0.502,0}{$\uparrow0.40$} & 0.3022          & 0.2736            & 0.2917             & 0.3004                 & \textcolor[rgb]{0,0.502,0}{$\downarrow0.0286$}  \\
\textit{+ Var-Both}                          & 63.97           & 63.90             & 64.33              & \textbf{64.37}                  & \textcolor[rgb]{0,0.502,0}{$\uparrow0.40$} & 0.3022          & \textbf{0.2500}            & 0.2853             & 0.3009                 & \textcolor[rgb]{0,0.502,0}{$\downarrow0.0522$}  \\
\Xhline{2\arrayrulewidth}
\end{tabular}
}
\caption{Full results of self-ensembling with different variations. We mark the cells of baseline systems without self-ensembling and their results in grey. Numbers in \textbf{bold} represents the best metric values for each learning method. $\Delta$ calculates the difference of performance and calibration error between the original results (Ori.) and the best self-ensembled results, where green means better results and red means worse results. We run all the experiments above 3 times if possible and show the mean values. }
\label{apptab:self ensembling full results}
\end{table*}

%% file: figs/latex/ece_plots.tex
\begin{figure*}[t!]
\centering
  \begin{subfigure}[b]{0.24\linewidth}
    \centering\includegraphics[width=\linewidth]{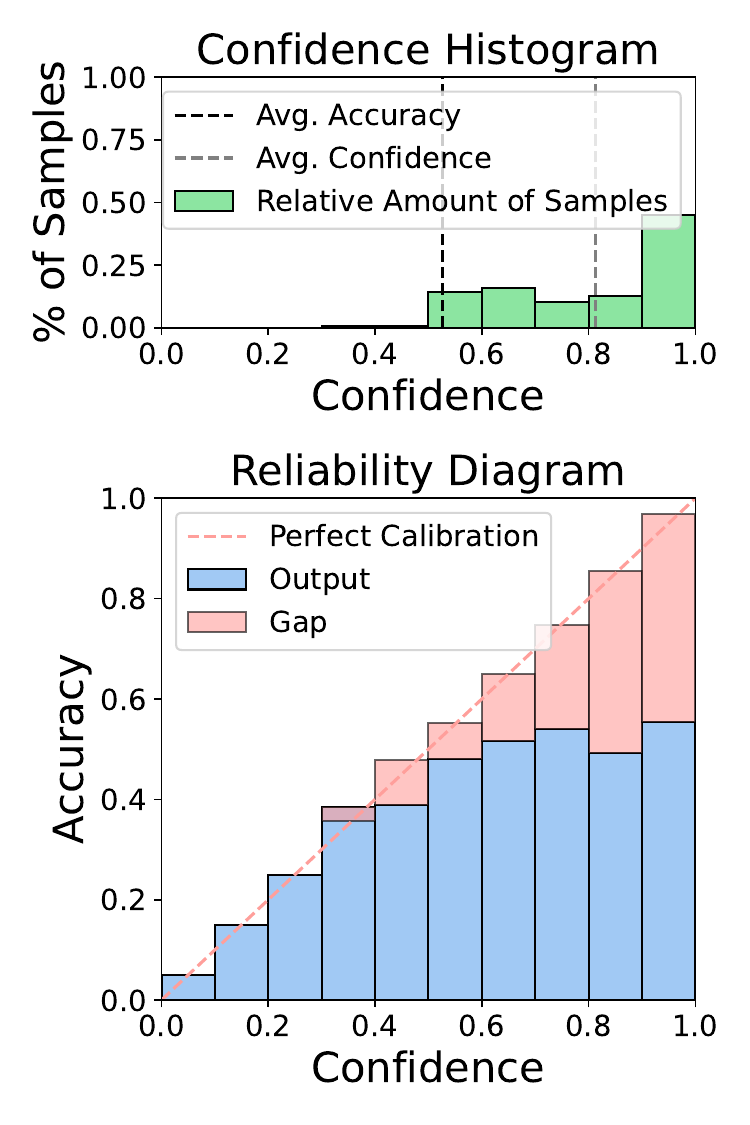}
    \caption{FT}
  \end{subfigure}
  \begin{subfigure}[b]{0.24\linewidth}
    \centering\includegraphics[width=\linewidth]{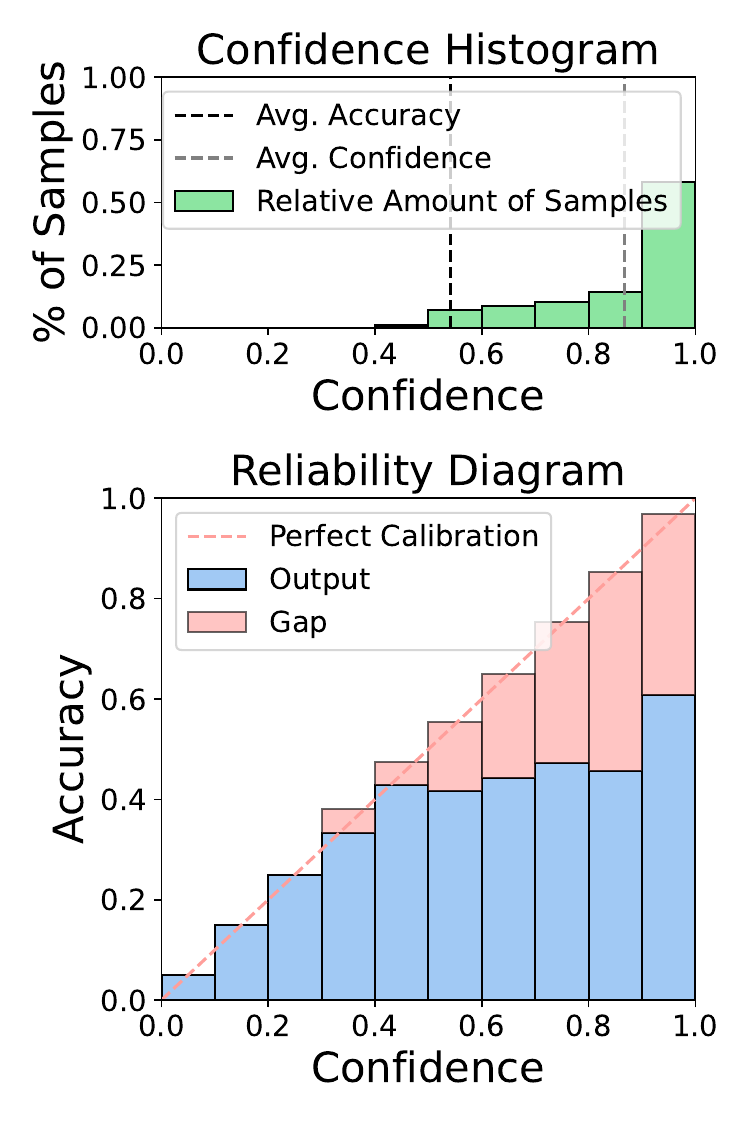}
    \caption{SICL}
  \end{subfigure}
  \begin{subfigure}[b]{0.24\linewidth}
    \centering\includegraphics[width=\linewidth]{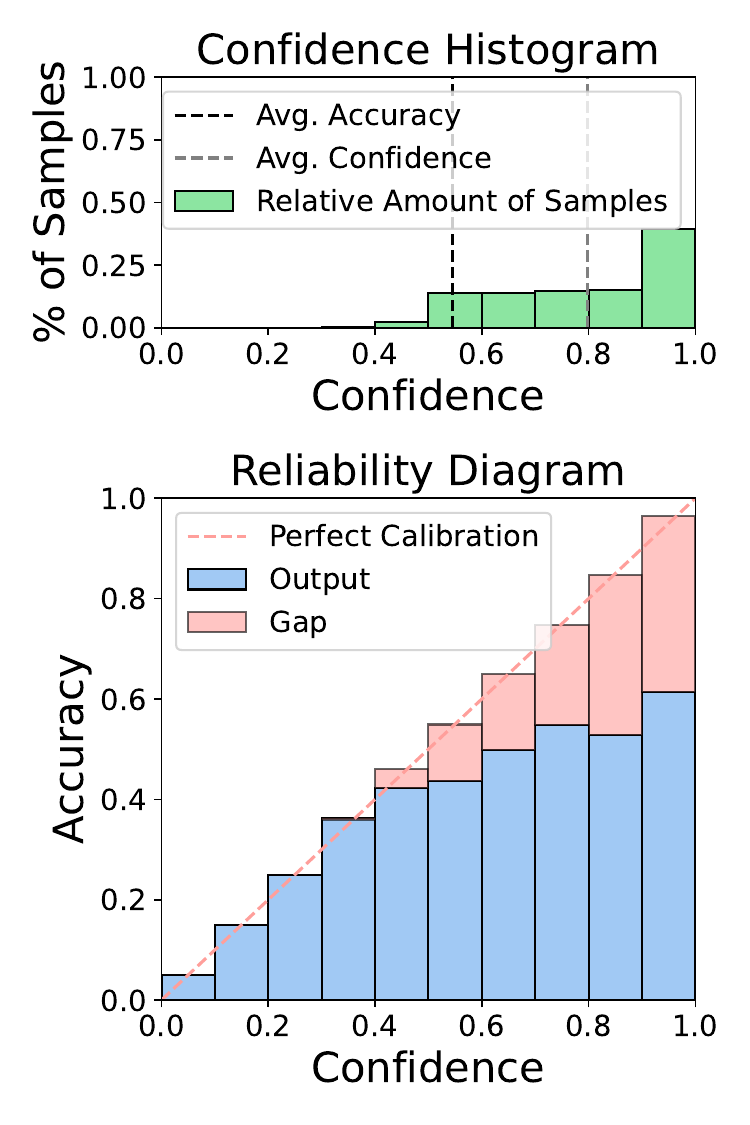}
    \caption{E-FT(\textit{Var-Prompt})}
  \end{subfigure}
  \begin{subfigure}[b]{0.24\linewidth}
    \centering\includegraphics[width=\linewidth]{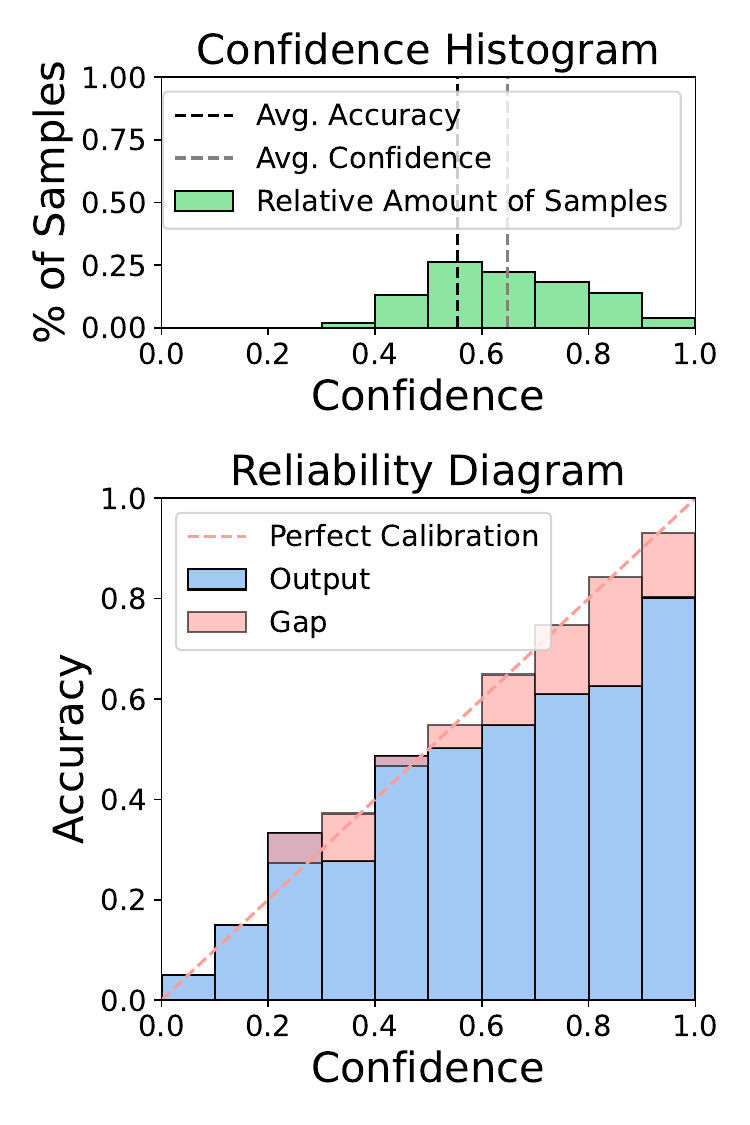}
    \caption{E-SICL(\textit{Var-Both})}
  \end{subfigure}
  \caption{The confidence histograms and reliability diagrams of SFT and SICL on SST-5 with or without self-ensembling using max probability. }
  \label{fig:eceplots}
\end{figure*}

%% file: appendix/further_ablations.tex
\section{Supplementary Results}
\label{sec:more ablations}

\subsection{Comparison with Classical Calibration Methods}
In comparison to self-ensembling in our work, both temperature scaling and Platt scaling \cite{guo2017calibration} require supervised labeled data as a validation set and parameter tuning (e.g., temperature) for different instruction templates, ICL settings, and models. In contrast, self-ensembling is tuning-free and employs strategies that are universally applicable across different learning methods. 
Furthermore, temperature scaling merely adjusts confidence levels while preserving the ranking among choices in classification tasks, making it ineffective in enhancing performance, as shown in Table \ref{apptab:comparison-temp-scaling}. 
Platt scaling is mostly suitable for binary classification tasks, making it less applicable to multi-class classification tasks.

We argue that self-ensembling is orthogonal to other calibration methods, allowing other calibration methods to be applied on top of the ensembled results for further calibration with improved performance. 
It can help mitigate the problem of miscalibration, offering new perspectives on calibration in addition to previous methods and enhancing trustworthiness.

Through experiments, we also observe that temperature scaling is sensitive to the selection of supervised labelled data. 
If we choose a batch of label-imbalanced supervised labelled data, temperature scaling may fail in this case. On the other hand, self-ensembling method doesn’t have this problem due to its training-free nature, as shown in Table \ref{apptab:imbalance-temp-scaling}.

\input{tables/appendix/temperature_scaling}

\subsection{How about larger models?}

Table \ref{apptab:xl main results} shows the results on SST-5 and Hate speech with different learning methods using Flan-T5\textsubscript{xl}. 
With ZSL and ICL, we observe that xl version model has larger calibration errors than Flan-T5\textsubscript{large} model on possibly seen datasets (SST-2 and SST-5), whereas on unseen datasets (Hate speech and Manifestos) it shows lower ECE.
Regarding the performances, the xl model shows better performances on unseen datasets than the large version model but doesn't guarantee better performances on seen datasets. 
After tuning the model with SFT or SICL, we find that the calibration errors are reduced across all tasks, which is different from Flan-T5\textsubscript{large}. 
Due to the computation constraint, we leave the discrepancy in the behaviors of different-sized models to future work.

Table \ref{apptab:xl ensemble results} shows the self-ensembled results using Flan-T5\textsubscript{xl} on SST-5 and Hate speech. 
We find that both the performances and calibration errors get better with self-ensembling, justifying the feasibility and extensibility of self-ensembling on larger models. 
Compared with Flan-T5\textsubscript{large}, the self-ensembled xl model yields much lower calibration errors with SFT and SICL on both tasks. 

We also surprisingly find that the self-ensembling method can improve both the performance and calibration on the task that Batch Calibration finds struggling. 
On Hate speech, after applying Batch Calibration, we witnessed an improvement in calibration along with a drop in performance. 
However, when we apply self-ensembling, the predictions yield better performance and much lower calibration errors at the same time.
This indicates the potential of self-ensembled language models in producing better and more reliable predictions.

\input{tables/appendix/xl_main_results_full}

\input{tables/appendix/xl_ensemble_results}

%% file: tables/appendix/temperature_scaling.tex
\begin{table*}
\centering
\renewcommand\arraystretch{1.2}
\resizebox{\textwidth}{!}{%
\begin{tabular}{l|cccccccc} 
\hline
\multicolumn{9}{c}{\textbf{SST-5 }}                                                                                                                                                                                                                               \\ 
\hline
                     & \multicolumn{2}{c}{ZSL}                                  & \multicolumn{2}{c}{ICL}                                  & \multicolumn{2}{c}{SFT}                                  & \multicolumn{2}{c}{SICL}                                  \\ 
\hline
                     & Performance & {\cellcolor[rgb]{0.937,0.937,0.937}}ECE    & Performance & {\cellcolor[rgb]{0.937,0.937,0.937}}ECE    & Performance & {\cellcolor[rgb]{0.937,0.937,0.937}}ECE    & Performance & {\cellcolor[rgb]{0.937,0.937,0.937}}ECE     \\
Original*            & 38.08       & {\cellcolor[rgb]{0.937,0.937,0.937}}0.2305 & 37.88       & {\cellcolor[rgb]{0.937,0.937,0.937}}0.1802 & 50.28       & {\cellcolor[rgb]{0.937,0.937,0.937}}0.4307 & 44.27       & {\cellcolor[rgb]{0.937,0.937,0.937}}0.3931  \\
Temperature Scaling* & 38.08       & {\cellcolor[rgb]{0.937,0.937,0.937}}0.0626 & 37.88       & {\cellcolor[rgb]{0.937,0.937,0.937}}0.0693 & 50.28       & {\cellcolor[rgb]{0.937,0.937,0.937}}0.0804 & 44.27       & {\cellcolor[rgb]{0.937,0.937,0.937}}0.0783  \\
Original             & 42.00       & {\cellcolor[rgb]{0.937,0.937,0.937}}0.1416 & 37.59       & {\cellcolor[rgb]{0.937,0.937,0.937}}0.1833 & 46.27       & {\cellcolor[rgb]{0.937,0.937,0.937}}0.403  & 47.12       & {\cellcolor[rgb]{0.937,0.937,0.937}}0.3602  \\
Self-Ensembling      & 46.79       & {\cellcolor[rgb]{0.937,0.937,0.937}}0.0782 & 37.75       & {\cellcolor[rgb]{0.937,0.937,0.937}}0.0955 & 48.35       & {\cellcolor[rgb]{0.937,0.937,0.937}}0.3008 & 47.91       & {\cellcolor[rgb]{0.937,0.937,0.937}}0.1200  \\ 
\hline
\multicolumn{9}{c}{\textbf{Hate Speech }}                                                                                                                                                                                                                         \\ 
\hline
                     & \multicolumn{2}{c}{ZSL}                                  & \multicolumn{2}{c}{ICL}                                  & \multicolumn{2}{c}{SFT}                                  & \multicolumn{2}{c}{SICL}                                  \\ 
\hline
                     & Performance & {\cellcolor[rgb]{0.937,0.937,0.937}}ECE    & Performance & {\cellcolor[rgb]{0.937,0.937,0.937}}ECE    & Performance & {\cellcolor[rgb]{0.937,0.937,0.937}}ECE    & Performance & {\cellcolor[rgb]{0.937,0.937,0.937}}ECE     \\
Original*            & 47.08       & {\cellcolor[rgb]{0.937,0.937,0.937}}0.2327 & 40.21       & {\cellcolor[rgb]{0.937,0.937,0.937}}0.2729 & 57.41       & {\cellcolor[rgb]{0.937,0.937,0.937}}0.3952 & 57.50       & {\cellcolor[rgb]{0.937,0.937,0.937}}0.2526  \\
Temperature Scaling* & 47.08       & {\cellcolor[rgb]{0.937,0.937,0.937}}0.0658 & 40.21       & {\cellcolor[rgb]{0.937,0.937,0.937}}0.0684 & 57.41       & {\cellcolor[rgb]{0.937,0.937,0.937}}0.0539 & 57.50       & {\cellcolor[rgb]{0.937,0.937,0.937}}0.0312  \\
Original             & 37.08       & {\cellcolor[rgb]{0.937,0.937,0.937}}0.3175 & 40.09       & {\cellcolor[rgb]{0.937,0.937,0.937}}0.2708 & 58.01       & {\cellcolor[rgb]{0.937,0.937,0.937}}0.3541 & 59.48       & {\cellcolor[rgb]{0.937,0.937,0.937}}0.1928  \\
Self-Ensembling      & 36.95       & {\cellcolor[rgb]{0.937,0.937,0.937}}0.2698 & 41.19       & {\cellcolor[rgb]{0.937,0.937,0.937}}0.1596 & 57.19       & {\cellcolor[rgb]{0.937,0.937,0.937}}0.2386 & 60.22       & {\cellcolor[rgb]{0.937,0.937,0.937}}0.1151  \\ 
\hline
\multicolumn{9}{c}{\textbf{Manifestos }}                                                                                                                                                                                                                          \\ 
\hline
                     & \multicolumn{2}{c}{ZSL}                                  & \multicolumn{2}{c}{ICL}                                  & \multicolumn{2}{c}{SFT}                                  & \multicolumn{2}{c}{SICL}                                  \\ 
\hline
                     & Performance & {\cellcolor[rgb]{0.937,0.937,0.937}}ECE    & Performance & {\cellcolor[rgb]{0.937,0.937,0.937}}ECE    & Performance & {\cellcolor[rgb]{0.937,0.937,0.937}}ECE    & Performance & {\cellcolor[rgb]{0.937,0.937,0.937}}ECE     \\
Original*            & 14.41       & {\cellcolor[rgb]{0.937,0.937,0.937}}0.4362 & 19.33       & {\cellcolor[rgb]{0.937,0.937,0.937}}0.4741 & 37.27       & {\cellcolor[rgb]{0.937,0.937,0.937}}0.2591 & 35.87       & {\cellcolor[rgb]{0.937,0.937,0.937}}0.2115  \\
Temperature Scaling* & 14.41       & {\cellcolor[rgb]{0.937,0.937,0.937}}0.0781 & 19.33       & {\cellcolor[rgb]{0.937,0.937,0.937}}0.0800 & 37.27       & {\cellcolor[rgb]{0.937,0.937,0.937}}0.0630 & 35.87       & {\cellcolor[rgb]{0.937,0.937,0.937}}0.0930  \\
Original             & 14.50       & {\cellcolor[rgb]{0.937,0.937,0.937}}0.4319 & 13.01       & {\cellcolor[rgb]{0.937,0.937,0.937}}0.4760 & 35.76       & {\cellcolor[rgb]{0.937,0.937,0.937}}0.2005 & 37.55       & {\cellcolor[rgb]{0.937,0.937,0.937}}0.2138  \\
Self-Ensembling      & 15.29       & {\cellcolor[rgb]{0.937,0.937,0.937}}0.2620 & 13.69       & {\cellcolor[rgb]{0.937,0.937,0.937}}0.1932 & 36.49       & {\cellcolor[rgb]{0.937,0.937,0.937}}0.0662 & 37.61       & {\cellcolor[rgb]{0.937,0.937,0.937}}0.1236  \\
\hline
\end{tabular}
}
\caption{Comparison with Temperature Scaling and Self-Ensembling. Self-ensembling helps to improve both performance and calibration where temperature scaling falls short in improving performance, making it a different perspective in calibration compared to temperature scaling. * indicates that the experiment results are only reported with 1 random seed. }
\label{apptab:comparison-temp-scaling}
\end{table*}

\begin{table*}[]
\renewcommand\arraystretch{1.2}
\resizebox{\textwidth}{!}{%
\begin{tabular}{l|cc|cc|cc|cc}
\hline
                    & \multicolumn{2}{c|}{ZSL}                                 & \multicolumn{2}{c|}{ICL}                                 & \multicolumn{2}{c|}{SFT}                                 & \multicolumn{2}{c}{SICL}                                 \\ \hline
                    & Performance    & \cellcolor[HTML]{EFEFEF}ECE             & Performance    & \cellcolor[HTML]{EFEFEF}ECE             & Performance    & \cellcolor[HTML]{EFEFEF}ECE             & Performance    & \cellcolor[HTML]{EFEFEF}ECE             \\
\textit{Original*}   & \textit{14.41} & \cellcolor[HTML]{EFEFEF}\textit{0.4362} & \textit{19.33} & \cellcolor[HTML]{EFEFEF}\textit{0.4741} & \textit{37.27} & \cellcolor[HTML]{EFEFEF}\textit{0.2591} & \textit{35.87} & \cellcolor[HTML]{EFEFEF}\textit{0.2115} \\
Temperature Scaling* & 14.41          & \cellcolor[HTML]{EFEFEF}0.9333          & 19.33          & \cellcolor[HTML]{EFEFEF}0.9333          & 37.27          & \cellcolor[HTML]{EFEFEF}0.9333          & 35.87          & \cellcolor[HTML]{EFEFEF}0.9333          \\
\textit{Original}   & \textit{14.50} & \cellcolor[HTML]{EFEFEF}\textit{0.4319} & \textit{13.01} & \cellcolor[HTML]{EFEFEF}\textit{0.4760} & \textit{35.76} & \cellcolor[HTML]{EFEFEF}\textit{0.2005} & \textit{37.55} & \cellcolor[HTML]{EFEFEF}\textit{0.2138} \\
Self-Ensembling     & 15.29          & \cellcolor[HTML]{EFEFEF}0.2620          & 13.69          & \cellcolor[HTML]{EFEFEF}0.1932          & 36.49          & \cellcolor[HTML]{EFEFEF}0.0662          & 37.61          & \cellcolor[HTML]{EFEFEF}0.1236          \\ \hline
\end{tabular}%
}
\caption{Temperature scaling is sensitive to the selection of supervised labelled data. When choosing a batch of label-imbalanced supervised labelled data from Manifestos, temperature scaling may fail in this case. On the other hand, self-ensembling method doesn’t have this problem due to its training-free nature. * indicates that the experiment results are only reported with 1 random seed. }
\label{apptab:imbalance-temp-scaling}
\end{table*}

%% file: tables/appendix/xl_main_results_full.tex
\begin{table*}
\centering
\renewcommand\arraystretch{1.3}
\resizebox{\linewidth}{!}{%
\begin{tabular}{ll|cccc|cccc} 
\Xhline{2\arrayrulewidth}
\multicolumn{2}{l|}{\multirow{2}{*}{\textbf{Evaluation Metrics }}}                                                                                   & \multicolumn{4}{c|}{\textbf{SST2 }}                                                                       & \multicolumn{4}{c}{\textbf{SST5 }}                                                                         \\
\multicolumn{2}{l|}{}                                                                                                                                & \textbf{ZSL} & \textbf{ICL}                 & \textbf{SFT}                  & \textbf{SICL}              & \textbf{ZSL} & \textbf{ICL}                 & \textbf{SFT}                  & \textbf{SICL}               \\ 
\hline
\multirow{2}{*}{\textbf{Performance }}                                                                                  & \textbf{acc}               & 96.38        & 96.81\textsubscript{0.12}    & 96.81\textsubscript{0.04}    & 96.89\textsubscript{0.05}    & 52.76        & 50.59\textsubscript{0.13}    & 52.93\textsubscript{0.28}    & 53.47\textsubscript{0.87}     \\
                                                                                                                        & \textbf{macro f1}          & -            & -                            & -                            & -                            & 38.01        & 31.28\textsubscript{0.32}    & 44.65\textsubscript{0.58}    & 43.91\textsubscript{1.04}     \\
\rowcolor[rgb]{0.900,0.900,0.900} {\cellcolor[rgb]{0.900,0.900,0.900}}                                                  & \textit{\textbf{acc}}      & 96.87        & 97.00\textsubscript{0.03}    & 96.78\textsubscript{0.05}    & 96.92\textsubscript{0.08}    & 50.59        & 52.01\textsubscript{0.15}    & 51.83\textsubscript{0.26}    & 51.95\textsubscript{0.04}     \\
\rowcolor[rgb]{0.900,0.900,0.900} \multirow{-2}{*}{{\cellcolor[rgb]{0.900,0.900,0.900}}\textbf{\textit{+ calibrated }}} & \textit{\textbf{macro f1}} & -            & -                            & -                            & -                            & 49.33        & 50.85\textsubscript{0.18}    & 51.00\textsubscript{0.38}    & 51.07\textsubscript{0.08}     \\
\textbf{Trustworthiness}                                                                                                & \textbf{ECE}               & 0.9291       & 0.9369\textsubscript{0.0001} & 0.9340\textsubscript{0.0011} & 0.9380\textsubscript{0.0003} & 0.2309       & 0.3065\textsubscript{0.0018} & 0.1689\textsubscript{0.0093} & 0.1744\textsubscript{0.0076}  \\
\rowcolor[rgb]{0.900,0.900,0.900} \textbf{\textit{+ calibrated}}                                                        & \textit{\textbf{ECE}}      & 0.7937       & 0.8000\textsubscript{0.0002} & 0.7930\textsubscript{0.0001} & 0.7988\textsubscript{0.0008} & 0.0925       & 0.1117\textsubscript{0.0014} & 0.1070\textsubscript{0.0049} & 0.1067\textsubscript{0.0014}  \\ 
\hline
\multicolumn{2}{l|}{\multirow{2}{*}{\textbf{Evaluation Metrics }}}                                                                                   & \multicolumn{4}{c|}{\textbf{Manifestos }}                                                                 & \multicolumn{4}{c}{\textbf{Hate Speech }}                                                                  \\
\multicolumn{2}{l|}{}                                                                                                                                & \textbf{ZSL} & \textbf{ICL}                 & \textbf{SFT}                  & \textbf{SICL}              & \textbf{ZSL} & \textbf{ICL}                 & \textbf{SFT}                  & \textbf{SICL}               \\ 
\hline
\multirow{2}{*}{\textbf{Performance }}                                                                                  & \textbf{micro f1}          & 28.37        & 31.37\textsubscript{0.18}    & 37.04\textsubscript{0.29}    & 38.25\textsubscript{0.44}    & 54.33        & 52.84\textsubscript{0.06}    & 64.44\textsubscript{0.51}    & 62.78\textsubscript{0.58}     \\
                                                                                                                        & \textbf{macro f1}          & 21.71        & 25.35\textsubscript{0.22}    & 33.71\textsubscript{0.62}    & 35.55\textsubscript{0.41}    & 47.15        & 46.49\textsubscript{0.54}    & 61.72\textsubscript{0.40}    & 60.27\textsubscript{0.31}     \\
\rowcolor[rgb]{0.900,0.900,0.900} {\cellcolor[rgb]{0.900,0.900,0.900}}                                                  & \textit{\textbf{micro f1}} & 37.38        & 37.96\textsubscript{0.36}    & 38.75\textsubscript{0.35}    & 38.83\textsubscript{0.85}    & 46.60        & 47.49\textsubscript{0.33}    & 61.07\textsubscript{2.57}    & 56.36\textsubscript{0.64}     \\
\rowcolor[rgb]{0.900,0.900,0.900} \multirow{-2}{*}{{\cellcolor[rgb]{0.900,0.900,0.900}}\textbf{\textit{+ calibrated }}} & \textit{\textbf{macro f1}} & 35.26        & 35.99\textsubscript{0.35}    & 36.17\textsubscript{0.37}    & 36.96\textsubscript{0.59}    & 43.65        & 44.78\textsubscript{0.35}    & 59.23\textsubscript{2.38}    & 54.80\textsubscript{0.66}     \\
\textbf{Trustworthiness}                                                                                                & \textbf{ECE}               & 0.4514       & 0.3839\textsubscript{0.0027} & 0.1116\textsubscript{0.0258} & 0.1066\textsubscript{0.0153} & 0.2309       & 0.1935\textsubscript{0.0015} & 0.1047\textsubscript{0.0446} & 0.0713\textsubscript{0.0063}  \\
\rowcolor[rgb]{0.900,0.900,0.900} \textbf{\textit{+ calibrated}}                                                        & \textit{\textbf{ECE}}      & 0.0402       & 0.0382\textsubscript{0.0067} & 0.0447\textsubscript{0.0063} & 0.0335\textsubscript{0.0070} & 0.0956       & 0.1076\textsubscript{0.0017} & 0.0297\textsubscript{0.0127} & 0.0474\textsubscript{0.0040}  \\
\Xhline{2\arrayrulewidth}
\end{tabular}
}
\caption{Experiment results with different learning methods using FlanT5-xl. We report the mean value for 3 runs with different random seeds and list the variance in the subscripts. We color the Batch Calibration results in grey. }
\label{apptab:xl main results}
\end{table*}

%% file: tables/appendix/xl_ensemble_results.tex
\begin{table*}
\centering
\arrayrulecolor{black}
\resizebox{\linewidth}{!}{%
\begin{tabular}{l|ccccc|ccccc} 
\Xhline{2\arrayrulewidth}
\multirow{3}{*}{\textbf{Systems }}       & \multicolumn{10}{c}{\textbf{SST5 }}                                                                                                                                                                                                                                      \\
                                         & \multicolumn{5}{c}{\textbf{Macro F1 }}                                                                                           & \multicolumn{5}{c}{\textbf{ECE }}                                                                                                     \\
                                         & \textbf{Ori.} & \textbf{Max} & \textbf{Mean} & \textbf{Majority} & \multicolumn{1}{c}{$\Delta$}                    & \textbf{Ori.} & \textbf{Max} & \textbf{Mean} & \textbf{Majority} & $\Delta$                                             \\ 
\hline
\rowcolor[rgb]{0.900,0.900,0.900} ZSL    & 38.01           &                   &                    &                        & \textcolor[rgb]{0,0.502,0}{$\uparrow5.53$} & 0.2309          &                   &                    &                        & \textcolor[rgb]{0,0.502,0}{$\downarrow0.0898$}  \\
\textit{+ Var-Prompt}                & 38.01           & \textbf{43.54}             & 41.80              & 42.65                  & \textcolor[rgb]{0,0.502,0}{$\uparrow5.53$} & 0.2309          & \textbf{0.1411}            & 0.1971             & 0.2210                 & \textcolor[rgb]{0,0.502,0}{$\downarrow0.0898$}  \\
\rowcolor[rgb]{0.900,0.900,0.900} ICL    & 31.28           &                   &                    &                        & \textcolor[rgb]{0,0.502,0}{$\uparrow3.95$} & 0.3065          &                   &                    &                        & \textcolor[rgb]{0,0.502,0}{$\downarrow0.1171$}  \\
\textit{+ Var-IC}              & 31.28           & 30.57             & 30.85              & 30.91                  & \textcolor{red}{$\downarrow0.37$}          & 0.3065          & 0.2653            & 0.3035             & 0.3070                 & \textcolor[rgb]{0,0.502,0}{$\downarrow0.0412$}  \\
\textit{+ Var-Prompt}                & 31.28           & 34.75             & 34.71              & 34.77                  & \textcolor[rgb]{0,0.502,0}{$\uparrow3.49$} & 0.3065          & 0.2358            & 0.2714             & 0.2798                 & \textcolor[rgb]{0,0.502,0}{$\downarrow0.0707$}  \\
\textit{+ Var-Both}                          & 31.28           & \textbf{35.23}             & 33.75              & 33.51                  & \textcolor[rgb]{0,0.502,0}{$\uparrow3.95$} & 0.3065          & \textbf{0.1894}            & 0.2730              & 0.2846                 & \textcolor[rgb]{0,0.502,0}{$\downarrow0.1171$}  \\
\rowcolor[rgb]{0.900,0.900,0.900} FT     & 44.65           &                   &                    &                        & \textcolor[rgb]{0,0.502,0}{$\uparrow3.33$} & 0.1689          &                   &                    &                        & \textcolor[rgb]{0,0.502,0}{$\downarrow0.0730$}  \\
\textit{+ Var-Prompt}                & 47.06           & \textbf{47.98}             & 47.15              & 47.01                  & \textcolor[rgb]{0,0.502,0}{$\uparrow0.92$} & 0.1656          & \textbf{0.0959}            & 0.1425             & 0.1660                 & \textcolor[rgb]{0,0.502,0}{$\downarrow0.0697$}  \\
\rowcolor[rgb]{0.900,0.900,0.900} SupICL & 43.91           &                   &                    &                        & \textcolor[rgb]{0,0.502,0}{$\uparrow1.26$} & 0.1744          &                   &                    &                        & \textcolor[rgb]{0,0.502,0}{$\downarrow0.0740$}  \\
\textit{+ Var-IC}              & 43.91           & 43.91             & 44.05              & 44.00                  & \textcolor[rgb]{0,0.502,0}{$\uparrow0.14$} & 0.1744          & 0.1376            & 0.1738             & 0.1742                 & \textcolor[rgb]{0,0.502,0}{$\downarrow0.0368$}  \\
\textit{+ Var-Prompt}                & 44.21           & 44.79             & 44.69              & 44.78                  & \textcolor[rgb]{0,0.502,0}{$\uparrow0.58$} & 0.1688          & 0.1278            & 0.1536             & 0.1666                 & \textcolor[rgb]{0,0.502,0}{$\downarrow0.0410$}  \\
\textit{+ Var-Both}                          & 44.21           & \textbf{45.17}             & 44.60              & 44.41                  & \textcolor[rgb]{0,0.502,0}{$\uparrow0.96$} & 0.1688          & \textbf{0.1004}            & 0.1488             & 0.1638                 & \textcolor[rgb]{0,0.502,0}{$\downarrow0.0684$}  \\ 
\hline
\multirow{3}{*}{\textbf{Systems }}       & \multicolumn{10}{c}{\textbf{Hate Speech }}                                                                                                                                                                                                                               \\
                                         & \multicolumn{5}{c}{\textbf{Macro F1 }}                                                                                           & \multicolumn{5}{c}{\textbf{ECE }}                                                                                                     \\
                                         & \textbf{Ori.} & \textbf{Max} & \textbf{Mean} & \textbf{Majority} & \multicolumn{1}{c}{$\Delta$}                    & \textbf{Ori.} & \textbf{Max} & \textbf{Mean} & \textbf{Majority} & $\Delta$                                             \\ 
\hline
\rowcolor[rgb]{0.900,0.900,0.900} ZSL    & 47.15           &                   &                    &                        & \textcolor[rgb]{0,0.502,0}{$\uparrow0.59$} & 0.2309          &                   &                    &                        & \textcolor[rgb]{0,0.502,0}{$\downarrow0.0508$}  \\
\textit{+ Var-Prompt}                & 47.15           & 46.57             & 47.30              & \textbf{47.74}                  & \textcolor[rgb]{0,0.502,0}{$\uparrow0.59$} & 0.2309          & \textbf{0.1801}            & 0.2041             & 0.2194                 & \textcolor[rgb]{0,0.502,0}{$\downarrow0.0508$}  \\
\rowcolor[rgb]{0.900,0.900,0.900} ICL    & 46.49           &                   &                    &                        & \textcolor[rgb]{0,0.502,0}{$\uparrow1.09$} & 0.1935          &                   &                    &                        & \textcolor[rgb]{0,0.502,0}{$\downarrow0.1052$}  \\
\textit{+ Var-IC}              & 46.49           & 47.03             & 46.94              & 46.74                  & \textcolor[rgb]{0,0.502,0}{$\uparrow0.54$} & 0.1935          & 0.1301            & 0.1800               & 0.1901                 & \textcolor[rgb]{0,0.502,0}{$\downarrow0.0634$}  \\
\textit{+ Var-Prompt}                & 46.49           & 47.44             & 47.06              & 47.57                  & \textcolor[rgb]{0,0.502,0}{$\uparrow1.08$} & 0.1935          & 0.1257            & 0.1625             & 0.1836                 & \textcolor[rgb]{0,0.502,0}{$\downarrow0.0678$}  \\
\textit{+ Var-Both}                          & 46.49           & \textbf{47.45}             & 46.80              & 46.68                  & \textcolor[rgb]{0,0.502,0}{$\uparrow1.09$} & 0.1935          & \textbf{0.0883}            & 0.1581             & 0.1867                 & \textcolor[rgb]{0,0.502,0}{$\downarrow0.1052$}  \\
\rowcolor[rgb]{0.900,0.900,0.900} FT     & 61.72           &                   &                    &                        & \textcolor{red}{$\downarrow0.72$}          & 0.1074          &                   &                    &                        & \textcolor[rgb]{0,0.502,0}{$\downarrow0.0572$}  \\
\textit{+ Var-Prompt}                & 60.04           & 60.71             & \textbf{61.00}              & 60.86                  & \textcolor[rgb]{0,0.502,0}{$\uparrow0.96$} & 0.0873          & \textbf{0.0502}            & 0.0719             & 0.0887                 & \textcolor[rgb]{0,0.502,0}{$\downarrow0.0371$}  \\
\rowcolor[rgb]{0.900,0.900,0.900} SupICL & 60.27           &                   &                    &                        & \textcolor[rgb]{0,0.502,0}{$\uparrow1.84$} & 0.0713          &                   &                    &                        & \textcolor[rgb]{0,0.502,0}{$\downarrow0.0437$}  \\
\textit{+ Var-IC}              & 60.27           & 60.88             & 60.72              & 60.68                  & \textcolor[rgb]{0,0.502,0}{$\uparrow0.61$} & 0.0713          & 0.0424            & 0.0636             & 0.0729                 & \textcolor[rgb]{0,0.502,0}{$\downarrow0.0289$}  \\
\textit{+ Var-Prompt}                & 60.37           & 60.97             & 61.50              & 61.45                  & \textcolor[rgb]{0,0.502,0}{$\uparrow1.13$} & 0.0883          & 0.0427            & 0.0649             & 0.0793                 & \textcolor[rgb]{0,0.502,0}{$\downarrow0.0456$}  \\
\textit{+ Var-Both}                          & 60.37           & 61.69             & \textbf{62.11}              & 61.74                  & \textcolor[rgb]{0,0.502,0}{$\uparrow1.74$} & 0.0833          & \textbf{0.0276}            & 0.0588             & 0.0816                 & \textcolor[rgb]{0,0.502,0}{$\downarrow0.0557$}  \\ 
\Xhline{2\arrayrulewidth}
\end{tabular}
}
\caption{Results of self-ensembling with different variations using FlanT5-xl. For simplicity, we omit the variance of 3 runs and only show the mean values. The notations follow previous patterns.}
\label{apptab:xl ensemble results}
\end{table*}

%% file: appendix/prompting_templates.tex
\section{Prompting templates}
\label{app:prompt template}

\subsection{Prompting templates for main experiments}
\label{app:prompt template main}
We provide the prompting templates for different datasets when comparing learning methods as follows. 

\paragraph{SST-2}
\begin{verbatim}
Classify this sentence's sentiment into 
'positive' or 'negative': <SENTENCE>
<LABEL>
\end{verbatim}

\paragraph{RTE}
\begin{verbatim}
Does Sentence1 entails Sentence2?
Sentence1: <SENTENCE1>
SENTENCE2: <SENTENCE2>
<LABEL>
\end{verbatim}

\paragraph{ANLI}
\begin{verbatim}
Does the premise entails the hypothesis? 
Premise: <PREMISE>
Hypothesis: <HYPOTHESIS>
<LABEL>
\end{verbatim}

\paragraph{SST-5}
\begin{verbatim}
Classify this sentence's sentiment into 
"terrible", "bad", "neutral", "good" or 
"great": <SENTENCE>
<LABEL>
\end{verbatim}

\paragraph{NLU++}
\begin{verbatim}
Here is a sentence: '<SENTENCE>'
Try to answer this question if possible 
with 'yes' or 'no': '<QUESTION>'
<LABEL>
\end{verbatim}

\paragraph{Manifestos}
\begin{verbatim}
Which category about US society does the 
sentence belong to from "other", "external 
relations", "freedom and democracy", 
"political system", "economy", "welfare 
and quality of life", "fabric of society", 
"social groups": <SENTENCE>
<LABEL>
\end{verbatim}

\paragraph{Hate Speech}
\begin{verbatim}
Classify this sentence's sentiment into 
"hate", "neutral" or "support": <SENTENCE>
<LABEL>
\end{verbatim}

\subsection{Prompting templates for \textit{Var-Prompt}}
\label{app:prompt template cycling}
Below show the various templates for prompt cycling. 
All the prompting templates are manually written without specific crafting. 

\paragraph{ANLI}
\begin{verbatim}
Does the premise entails the hypothesis? 
Premise: <PREMISE>
Hypothesis: <HYPOTHESIS>
<LABEL>

Premise: <PREMISE>
Hypothesis: <HYPOTHESIS>
Given the premise, is the hypothesis 
entailed? 
<LABEL>

Is the hypothesis entailed by the premise? 
Premise: <PREMISE>
Hypothesis: <HYPOTHESIS>
<LABEL>

### Instruction: Determine whether the 
hypothesis is entailed by the premise. 
Premise: <PREMISE>
Hypothesis: <HYPOTHESIS>
<LABEL>
\end{verbatim}

\paragraph{SST-5}
\begin{verbatim}
Classify this sentence's sentiment into 
"terrible", "bad", "neutral", "good" or 
"great": <SENTENCE>
<LABEL>

<SENTENCE>
is this sentence 'great', 'good', 'neutral', 
'bad' or 'terrible'? 
<LABEL>

<SENTENCE>
Among "terrible", "bad", "neutral", "good"
or "great", the sentence's sentiment is 
<LABEL>

### Instruction: Classify the input 
sentence's sentiment into into "terrible", 
"bad", "neutral", "good" or "great". 
Input: <SENTENCE> 
### Response: <LABEL>
\end{verbatim}

\paragraph{Manifestos}
\begin{verbatim}
Which category about US society does the 
sentence belong to from "other", "external 
relations", "freedom and democracy", 
"political system", "economy", "welfare 
and quality of life", "fabric of society", 
"social groups": <SENTENCE>
<LABEL>

<SENTENCE>
Which category about US society does 
the sentence belong to?
<LABEL>

<SENTENCE>
Among "other", "external", "democracy", 
"political", "economy", "welfare", 
"fabric", "group", the sentence's US 
societal category is <LABEL>

### Instruction: Classify the input 
sentence's US societal category into 
"other", "external", "democracy", 
"political", "economy", "welfare", 
"fabric", "group". 
Input: <SENTENCE> 
### Response: <LABEL>
\end{verbatim}

\paragraph{Hate Speech}
\begin{verbatim}
Classify this sentence's sentiment into 
"hate", "neutral" or "support": <SENTENCE>
<LABEL>

<SENTENCE>
Is the sentence hate, neutral or support? 
<LABEL>

<SENTENCE>
Among "hate", "neutral" or "support", the 
sentence's sentiment is <LABEL>

### Instruction: What's the sentiment of 
input sentence among "hate", "neutral" 
or "support"? 
Input: <SENTENCE> 
### Response: <LABEL>
\end{verbatim}

\subsection{Prompting templates for \textit{Var-Prompt} in ablation studies}
\label{app:prompt ablation}
We use ChatGPT to generate paraphrased prompting templates for \textit{Var-Prompt}. 
The instruction we give to ChatGPT is as follows. 
\begin{verbatim}
Paraphrase the provided templates and keep 
the keywords in <> in the meantime. Show me 
5 different paraphrased results. 

The template is: 
<TEMPLATE>
\end{verbatim}

We paraphrase the prompting templates for SST-5 and Hate Speech datasets and randomly sampled 4 paraphrased candidates.
These templates share similar wording and structure with the human-written templates. 
We conduct experiments with these 8 templates in total and use max probability when self-ensembling. 
We provide the candidates below for reference. 

\paragraph{SST-5}

\begin{verbatim}
<SENTENCE>\nThe sentiment expressed by 
<SENTENCE> falls into the categories of 
"terrible," "bad," "neutral," "good," 
or "great," and it is labeled as <LABEL>

Evaluate the sentiment expressed by 
<SENTENCE>, placing it in the categories 
of "terrible," "bad," "neutral," "good," 
or "great," and indicate the sentiment 
as <LABEL>

Evaluate the emotional tone of this statement 
and categorize it as "terrible," "bad," 
"neutral," "good," or "great": <SENTENCE>
<LABEL>

Analyze the emotional inclination of the 
following statement, categorizing it as 
"terrible," "bad," "neutral," "good," or 
"great": <SENTENCE>
<LABEL>
\end{verbatim}

\paragraph{Hate Speech}
\begin{verbatim}
Assess whether the sentiment in this 
sentence falls under "hate," "neutral," 
or "support": <SENTENCE>
<LABEL>

Appraise the sentiment expressed in this 
sentence and assign it to one of the 
categories: "hate," "neutral," or "support"
: <SENTENCE>
<LABEL>

Categorize the sentiment of <SENTENCE> as 
either "hate," "neutral," or "support," 
with the assigned label being <LABEL>.

Determine the emotional tone of <SENTENCE>, 
categorizing it as "hate," "neutral," or 
"support," and mark the sentiment as <LABEL>.
\end{verbatim}

%% file: acl.bbl
\begin{thebibliography}{54}
\expandafter\ifx\csname natexlab\endcsname\relax\def\natexlab#1{#1}\fi

\bibitem[{Brown et~al.(2020)Brown, Mann, Ryder, Subbiah, Kaplan, Dhariwal, Neelakantan, Shyam, Sastry, Askell et~al.}]{brown2020language}
Tom Brown, Benjamin Mann, Nick Ryder, Melanie Subbiah, Jared~D Kaplan, Prafulla Dhariwal, Arvind Neelakantan, Pranav Shyam, Girish Sastry, Amanda Askell, et~al. 2020.
\newblock \href {https://proceedings.neurips.cc/paper/2020/hash/1457c0d6bfcb4967418bfb8ac142f64a-Abstract.html} {Language models are few-shot learners}.
\newblock \emph{Advances in neural information processing systems}, 33:1877--1901.

\bibitem[{Casanueva et~al.(2022)Casanueva, Vuli{\'c}, Spithourakis, and Budzianowski}]{casanueva-etal-2022-nlu}
Inigo Casanueva, Ivan Vuli{\'c}, Georgios Spithourakis, and Pawe{\l} Budzianowski. 2022.
\newblock \href {https://doi.org/10.18653/v1/2022.findings-naacl.154} {{NLU}++: A multi-label, slot-rich, generalisable dataset for natural language understanding in task-oriented dialogue}.
\newblock In \emph{Findings of the Association for Computational Linguistics: NAACL 2022}, pages 1998--2013, Seattle, United States. Association for Computational Linguistics.

\bibitem[{Chen et~al.(2022)Chen, Zhong, Zha, Karypis, and He}]{chen-etal-2022-meta}
Yanda Chen, Ruiqi Zhong, Sheng Zha, George Karypis, and He~He. 2022.
\newblock \href {https://doi.org/10.18653/v1/2022.acl-long.53} {Meta-learning via language model in-context tuning}.
\newblock In \emph{Proceedings of the 60th Annual Meeting of the Association for Computational Linguistics (Volume 1: Long Papers)}, pages 719--730, Dublin, Ireland. Association for Computational Linguistics.

\bibitem[{Chowdhery et~al.(2022)Chowdhery, Narang, Devlin, Bosma, Mishra, Roberts, Barham, Chung, Sutton, Gehrmann, Schuh, Shi, Tsvyashchenko, Maynez, Rao, Barnes, Tay, Shazeer, Prabhakaran, Reif, Du, Hutchinson, Pope, Bradbury, Austin, Isard, Gur-Ari, Yin, Duke, Levskaya, Ghemawat, Dev, Michalewski, Garc{\'i}a, Misra, Robinson, Fedus, Zhou, Ippolito, Luan, Lim, Zoph, Spiridonov, Sepassi, Dohan, Agrawal, Omernick, Dai, Pillai, Pellat, Lewkowycz, Moreira, Child, Polozov, Lee, Zhou, Wang, Saeta, D{\'i}az, Firat, Catasta, Wei, Meier-Hellstern, Eck, Dean, Petrov, and Fiedel}]{Chowdhery2022PaLMSL}
Aakanksha Chowdhery, Sharan Narang, Jacob Devlin, Maarten Bosma, Gaurav Mishra, Adam Roberts, Paul Barham, Hyung~Won Chung, Charles Sutton, Sebastian Gehrmann, Parker Schuh, Kensen Shi, Sasha Tsvyashchenko, Joshua Maynez, Abhishek Rao, Parker Barnes, Yi~Tay, Noam~M. Shazeer, Vinodkumar Prabhakaran, Emily Reif, Nan Du, Benton~C. Hutchinson, Reiner Pope, James Bradbury, Jacob Austin, Michael Isard, Guy Gur-Ari, Pengcheng Yin, Toju Duke, Anselm Levskaya, Sanjay Ghemawat, Sunipa Dev, Henryk Michalewski, Xavier Garc{\'i}a, Vedant Misra, Kevin Robinson, Liam Fedus, Denny Zhou, Daphne Ippolito, David Luan, Hyeontaek Lim, Barret Zoph, Alexander Spiridonov, Ryan Sepassi, David Dohan, Shivani Agrawal, Mark Omernick, Andrew~M. Dai, Thanumalayan~Sankaranarayana Pillai, Marie Pellat, Aitor Lewkowycz, Erica Moreira, Rewon Child, Oleksandr Polozov, Katherine Lee, Zongwei Zhou, Xuezhi Wang, Brennan Saeta, Mark D{\'i}az, Orhan Firat, Michele Catasta, Jason Wei, Kathleen~S. Meier-Hellstern, Douglas Eck, Jeff Dean, Slav Petrov,
  and Noah Fiedel. 2022.
\newblock \href {https://api.semanticscholar.org/CorpusID:247951931} {Palm: Scaling language modeling with pathways}.
\newblock \emph{J. Mach. Learn. Res.}, 24:240:1--240:113.

\bibitem[{Chung et~al.(2022)Chung, Hou, Longpre, Zoph, Tay, Fedus, Li, Wang, Dehghani, Brahma et~al.}]{chung2022scaling}
Hyung~Won Chung, Le~Hou, Shayne Longpre, Barret Zoph, Yi~Tay, William Fedus, Yunxuan Li, Xuezhi Wang, Mostafa Dehghani, Siddhartha Brahma, et~al. 2022.
\newblock \href {https://doi.org/10.48550/arXiv.2210.11416} {Scaling instruction-finetuned language models}.
\newblock \emph{arXiv preprint arXiv:2210.11416}.

\bibitem[{Deng et~al.(2023)Deng, Zhao, Tang, Gerstein, and Cohan}]{deng2023investigating}
Chunyuan Deng, Yilun Zhao, Xiangru Tang, Mark Gerstein, and Arman Cohan. 2023.
\newblock \href {https://doi.org/10.48550/arXiv.2311.09783} {Investigating data contamination in modern benchmarks for large language models}.
\newblock \emph{arXiv preprint arXiv:2311.09783}.

\bibitem[{Desai and Durrett(2020)}]{desai2020calibration}
Shrey Desai and Greg Durrett. 2020.
\newblock \href {https://doi.org/10.18653/v1/2020.emnlp-main.21} {Calibration of pre-trained transformers}.
\newblock In \emph{Proceedings of the 2020 Conference on Empirical Methods in Natural Language Processing (EMNLP)}, pages 295--302, Online. Association for Computational Linguistics.

\bibitem[{Devlin et~al.(2019)Devlin, Chang, Lee, and Toutanova}]{devlin-etal-2019-bert}
Jacob Devlin, Ming-Wei Chang, Kenton Lee, and Kristina Toutanova. 2019.
\newblock \href {https://doi.org/10.18653/v1/N19-1423} {{BERT}: Pre-training of deep bidirectional transformers for language understanding}.
\newblock In \emph{Proceedings of the 2019 Conference of the North {A}merican Chapter of the Association for Computational Linguistics: Human Language Technologies, Volume 1 (Long and Short Papers)}, pages 4171--4186, Minneapolis, Minnesota. Association for Computational Linguistics.

\bibitem[{Dong et~al.(2022)Dong, Li, Dai, Zheng, Wu, Chang, Sun, Xu, and Sui}]{dong2022survey}
Qingxiu Dong, Lei Li, Damai Dai, Ce~Zheng, Zhiyong Wu, Baobao Chang, Xu~Sun, Jingjing Xu, and Zhifang Sui. 2022.
\newblock \href {https://doi.org/10.48550/arXiv.2301.00234} {A survey for in-context learning}.
\newblock \emph{arXiv preprint arXiv:2301.00234}.

\bibitem[{Duan et~al.(2023)Duan, Tang, Yang, Abbasi, and Tam}]{duan2023exploring}
Hanyu Duan, Yixuan Tang, Yi~Yang, Ahmed Abbasi, and Kar~Yan Tam. 2023.
\newblock \href {https://doi.org/10.48550/arXiv.2311.10367} {Exploring the relationship between in-context learning and instruction tuning}.
\newblock \emph{arXiv preprint arXiv:2311.10367}.

\bibitem[{Fei et~al.(2023)Fei, Hou, Chen, and Bosselut}]{fei-etal-2023-mitigating}
Yu~Fei, Yifan Hou, Zeming Chen, and Antoine Bosselut. 2023.
\newblock \href {https://doi.org/10.18653/v1/2023.acl-long.783} {Mitigating label biases for in-context learning}.
\newblock In \emph{Proceedings of the 61st Annual Meeting of the Association for Computational Linguistics (Volume 1: Long Papers)}, pages 14014--14031, Toronto, Canada. Association for Computational Linguistics.

\bibitem[{Gleave and Irving(2022)}]{gleave2022uncertainty}
Adam Gleave and Geoffrey Irving. 2022.
\newblock \href {https://doi.org/10.48550/arXiv.2203.07472} {Uncertainty estimation for language reward models}.
\newblock \emph{arXiv preprint arXiv:2203.07472}.

\bibitem[{Gu et~al.(2023)Gu, Dong, Wei, and Huang}]{gu-etal-2023-pre}
Yuxian Gu, Li~Dong, Furu Wei, and Minlie Huang. 2023.
\newblock \href {https://doi.org/10.18653/v1/2023.acl-long.267} {Pre-training to learn in context}.
\newblock In \emph{Proceedings of the 61st Annual Meeting of the Association for Computational Linguistics (Volume 1: Long Papers)}, pages 4849--4870, Toronto, Canada. Association for Computational Linguistics.

\bibitem[{Guo et~al.(2017)Guo, Pleiss, Sun, and Weinberger}]{guo2017calibration}
Chuan Guo, Geoff Pleiss, Yu~Sun, and Kilian~Q Weinberger. 2017.
\newblock \href {http://proceedings.mlr.press/v70/guo17a.html} {On calibration of modern neural networks}.
\newblock In \emph{International conference on machine learning}, pages 1321--1330. PMLR.

\bibitem[{Hastie et~al.(2001)Hastie, Tibshirani, and Friedman}]{hastie01statisticallearning}
Trevor Hastie, Robert Tibshirani, and Jerome Friedman. 2001.
\newblock \emph{The Elements of Statistical Learning}.
\newblock Springer Series in Statistics. Springer New York Inc., New York, NY, USA.

\bibitem[{He et~al.(2022)He, Zhou, Ma, Berg{-}Kirkpatrick, and Neubig}]{he2021towards}
Junxian He, Chunting Zhou, Xuezhe Ma, Taylor Berg{-}Kirkpatrick, and Graham Neubig. 2022.
\newblock \href {https://openreview.net/forum?id=0RDcd5Axok} {Towards a unified view of parameter-efficient transfer learning}.
\newblock In \emph{The Tenth International Conference on Learning Representations, {ICLR} 2022, Virtual Event, April 25-29, 2022}. OpenReview.net.

\bibitem[{Hu et~al.(2022)Hu, Shen, Wallis, Allen{-}Zhu, Li, Wang, Wang, and Chen}]{hu2021lora}
Edward~J. Hu, Yelong Shen, Phillip Wallis, Zeyuan Allen{-}Zhu, Yuanzhi Li, Shean Wang, Lu~Wang, and Weizhu Chen. 2022.
\newblock \href {https://openreview.net/forum?id=nZeVKeeFYf9} {Lora: Low-rank adaptation of large language models}.
\newblock In \emph{The Tenth International Conference on Learning Representations, {ICLR} 2022, Virtual Event, April 25-29, 2022}. OpenReview.net.

\bibitem[{Jiang et~al.(2021)Jiang, Araki, Ding, and Neubig}]{jiang2021can}
Zhengbao Jiang, Jun Araki, Haibo Ding, and Graham Neubig. 2021.
\newblock \href {https://doi.org/10.1162/tacl_a_00407} {How can we know when language models know? on the calibration of language models for question answering}.
\newblock \emph{Transactions of the Association for Computational Linguistics}, 9:962--977.

\bibitem[{Kong et~al.(2020)Kong, Jiang, Zhuang, Lyu, Zhao, and Zhang}]{kong2020calibrated}
Lingkai Kong, Haoming Jiang, Yuchen Zhuang, Jie Lyu, Tuo Zhao, and Chao Zhang. 2020.
\newblock \href {https://doi.org/10.18653/v1/2020.emnlp-main.102} {Calibrated language model fine-tuning for in- and out-of-distribution data}.
\newblock In \emph{Proceedings of the 2020 Conference on Empirical Methods in Natural Language Processing (EMNLP)}, pages 1326--1340, Online. Association for Computational Linguistics.

\bibitem[{Lakshminarayanan et~al.(2017)Lakshminarayanan, Pritzel, and Blundell}]{lakshminarayanan2017simple}
Balaji Lakshminarayanan, Alexander Pritzel, and Charles Blundell. 2017.
\newblock \href {https://proceedings.neurips.cc/paper/2017/hash/9ef2ed4b7fd2c810847ffa5fa85bce38-Abstract.html} {Simple and scalable predictive uncertainty estimation using deep ensembles}.
\newblock \emph{Advances in neural information processing systems}, 30.

\bibitem[{Lehmann et~al.(2023)Lehmann, Franzmann, Burst, Regel, Riethmüller, Volkens, Weßels, and Zehnter}]{Lehmann:2023}
Pola Lehmann, Simon Franzmann, Tobias Burst, Sven Regel, Felicia Riethmüller, Andrea Volkens, Bernhard Weßels, and Lisa Zehnter. 2023.
\newblock \href {https://doi.org/10.25522/manifesto.mpds.2023a} {The manifesto data collection. manifesto project (mrg/cmp/marpor). version 2023a}.

\bibitem[{Liu et~al.(2024)Liu, Zhou, Guo, Shareghi, Vulic, Korhonen, and Collier}]{liu2024aligning}
Yinhong Liu, Han Zhou, Zhijiang Guo, Ehsan Shareghi, Ivan Vulic, Anna Korhonen, and Nigel Collier. 2024.
\newblock Aligning with human judgement: The role of pairwise preference in large language model evaluators.
\newblock \emph{arXiv preprint arXiv:2403.16950}.

\bibitem[{Longpre et~al.(2023)Longpre, Hou, Vu, Webson, Chung, Tay, Zhou, Le, Zoph, Wei, and Roberts}]{longpre2023flan}
Shayne Longpre, Le~Hou, Tu~Vu, Albert Webson, Hyung~Won Chung, Yi~Tay, Denny Zhou, Quoc~V. Le, Barret Zoph, Jason Wei, and Adam Roberts. 2023.
\newblock \href {https://proceedings.mlr.press/v202/longpre23a.html} {The flan collection: Designing data and methods for effective instruction tuning}.
\newblock In \emph{International Conference on Machine Learning, {ICML} 2023, 23-29 July 2023, Honolulu, Hawaii, {USA}}, volume 202 of \emph{Proceedings of Machine Learning Research}, pages 22631--22648. {PMLR}.

\bibitem[{Min et~al.(2022)Min, Lewis, Zettlemoyer, and Hajishirzi}]{min-etal-2022-metaicl}
Sewon Min, Mike Lewis, Luke Zettlemoyer, and Hannaneh Hajishirzi. 2022.
\newblock \href {https://doi.org/10.18653/v1/2022.naacl-main.201} {{M}eta{ICL}: Learning to learn in context}.
\newblock In \emph{Proceedings of the 2022 Conference of the North American Chapter of the Association for Computational Linguistics: Human Language Technologies}, pages 2791--2809, Seattle, United States. Association for Computational Linguistics.

\bibitem[{Mishra et~al.(2022)Mishra, Khashabi, Baral, Choi, and Hajishirzi}]{mishra2021reframing}
Swaroop Mishra, Daniel Khashabi, Chitta Baral, Yejin Choi, and Hannaneh Hajishirzi. 2022.
\newblock \href {https://doi.org/10.18653/v1/2022.findings-acl.50} {Reframing instructional prompts to {GPT}k{'}s language}.
\newblock In \emph{Findings of the Association for Computational Linguistics: ACL 2022}, pages 589--612, Dublin, Ireland. Association for Computational Linguistics.

\bibitem[{Mohammed and Kora(2023)}]{MOHAMMED2023757}
Ammar Mohammed and Rania Kora. 2023.
\newblock \href {https://doi.org/https://doi.org/10.1016/j.jksuci.2023.01.014} {A comprehensive review on ensemble deep learning: Opportunities and challenges}.
\newblock \emph{Journal of King Saud University - Computer and Information Sciences}, 35(2):757--774.

\bibitem[{Mosbach et~al.(2023)Mosbach, Pimentel, Ravfogel, Klakow, and Elazar}]{mosbach2023few}
Marius Mosbach, Tiago Pimentel, Shauli Ravfogel, Dietrich Klakow, and Yanai Elazar. 2023.
\newblock \href {https://doi.org/10.18653/v1/2023.findings-acl.779} {Few-shot fine-tuning vs. in-context learning: A fair comparison and evaluation}.
\newblock In \emph{Findings of the Association for Computational Linguistics: ACL 2023}, pages 12284--12314, Toronto, Canada. Association for Computational Linguistics.

\bibitem[{Nie et~al.(2020)Nie, Williams, Dinan, Bansal, Weston, and Kiela}]{nie-etal-2020-adversarial}
Yixin Nie, Adina Williams, Emily Dinan, Mohit Bansal, Jason Weston, and Douwe Kiela. 2020.
\newblock \href {https://doi.org/10.18653/v1/2020.acl-main.441} {Adversarial {NLI}: A new benchmark for natural language understanding}.
\newblock In \emph{Proceedings of the 58th Annual Meeting of the Association for Computational Linguistics}, pages 4885--4901, Online. Association for Computational Linguistics.

\bibitem[{OpenAI(2023)}]{OpenAI2023GPT4TR}
OpenAI. 2023.
\newblock \href {https://api.semanticscholar.org/CorpusID:257532815} {Gpt-4 technical report}.
\newblock \emph{ArXiv}, abs/2303.08774.

\bibitem[{Ovadia et~al.(2019)Ovadia, Fertig, Ren, Nado, Sculley, Nowozin, Dillon, Lakshminarayanan, and Snoek}]{ovadia2019can}
Yaniv Ovadia, Emily Fertig, Jie Ren, Zachary Nado, David Sculley, Sebastian Nowozin, Joshua Dillon, Balaji Lakshminarayanan, and Jasper Snoek. 2019.
\newblock \href {https://proceedings.neurips.cc/paper/2019/hash/8558cb408c1d76621371888657d2eb1d-Abstract.html} {Can you trust your model's uncertainty? evaluating predictive uncertainty under dataset shift}.
\newblock \emph{Advances in neural information processing systems}, 32.

\bibitem[{Radford et~al.(2019)Radford, Wu, Child, Luan, Amodei, and Sutskever}]{Radford2019LanguageMA}
Alec Radford, Jeff Wu, Rewon Child, David Luan, Dario Amodei, and Ilya Sutskever. 2019.
\newblock \href {https://api.semanticscholar.org/CorpusID:160025533} {Language models are unsupervised multitask learners}.

\bibitem[{Raffel et~al.(2020)Raffel, Shazeer, Roberts, Lee, Narang, Matena, Zhou, Li, and Liu}]{raffel2020exploring}
Colin Raffel, Noam Shazeer, Adam Roberts, Katherine Lee, Sharan Narang, Michael Matena, Yanqi Zhou, Wei Li, and Peter~J Liu. 2020.
\newblock \href {http://jmlr.org/papers/v21/20-074.html} {Exploring the limits of transfer learning with a unified text-to-text transformer}.
\newblock \emph{The Journal of Machine Learning Research}, 21(1):5485--5551.

\bibitem[{Razumovskaia et~al.(2023)Razumovskaia, Glava{\v{s}}, Korhonen, and Vuli{\'c}}]{razumovskaia2023sqatin}
Evgeniia Razumovskaia, Goran Glava{\v{s}}, Anna Korhonen, and Ivan Vuli{\'c}. 2023.
\newblock \href {https://doi.org/10.48550/arXiv.2311.09502} {Sqatin: Supervised instruction tuning meets question answering for improved dialogue nlu}.
\newblock \emph{arXiv preprint arXiv:2311.09502}.

\bibitem[{Sachdeva et~al.(2022)Sachdeva, Barreto, Bacon, Sahn, von Vacano, and Kennedy}]{sachdeva-etal-2022-measuring}
Pratik Sachdeva, Renata Barreto, Geoff Bacon, Alexander Sahn, Claudia von Vacano, and Chris Kennedy. 2022.
\newblock \href {https://aclanthology.org/2022.nlperspectives-1.11} {The measuring hate speech corpus: Leveraging rasch measurement theory for data perspectivism}.
\newblock In \emph{Proceedings of the 1st Workshop on Perspectivist Approaches to NLP @LREC2022}, pages 83--94, Marseille, France. European Language Resources Association.

\bibitem[{Shi et~al.(2023)Shi, Min, Lomeli, Zhou, Li, Lin, Smith, Zettlemoyer, Yih, and Lewis}]{shi2023context}
Weijia Shi, Sewon Min, Maria Lomeli, Chunting Zhou, Margaret Li, Victoria Lin, Noah~A Smith, Luke Zettlemoyer, Scott Yih, and Mike Lewis. 2023.
\newblock \href {https://doi.org/10.48550/arXiv.2310.10638} {In-context pretraining: Language modeling beyond document boundaries}.
\newblock \emph{arXiv preprint arXiv:2310.10638}.

\bibitem[{Singhal et~al.(2022)Singhal, Azizi, Tu, Mahdavi, Wei, Chung, Scales, Tanwani, Cole-Lewis, Pfohl et~al.}]{singhal2022large}
Karan Singhal, Shekoofeh Azizi, Tao Tu, S~Sara Mahdavi, Jason Wei, Hyung~Won Chung, Nathan Scales, Ajay Tanwani, Heather Cole-Lewis, Stephen Pfohl, et~al. 2022.
\newblock \href {https://doi.org/10.48550/arXiv.2212.13138} {Large language models encode clinical knowledge}.
\newblock \emph{arXiv preprint arXiv:2212.13138}.

\bibitem[{Socher et~al.(2013)Socher, Perelygin, Wu, Chuang, Manning, Ng, and Potts}]{socher-etal-2013-recursive}
Richard Socher, Alex Perelygin, Jean Wu, Jason Chuang, Christopher~D. Manning, Andrew Ng, and Christopher Potts. 2013.
\newblock \href {https://aclanthology.org/D13-1170} {Recursive deep models for semantic compositionality over a sentiment treebank}.
\newblock In \emph{Proceedings of the 2013 Conference on Empirical Methods in Natural Language Processing}, pages 1631--1642, Seattle, Washington, USA. Association for Computational Linguistics.

\bibitem[{Su et~al.(2023)Su, Kasai, Wu, Shi, Wang, Xin, Zhang, Ostendorf, Zettlemoyer, Smith, and Yu}]{su2022selective}
Hongjin Su, Jungo Kasai, Chen~Henry Wu, Weijia Shi, Tianlu Wang, Jiayi Xin, Rui Zhang, Mari Ostendorf, Luke Zettlemoyer, Noah~A. Smith, and Tao Yu. 2023.
\newblock \href {https://openreview.net/pdf?id=qY1hlv7gwg} {Selective annotation makes language models better few-shot learners}.
\newblock In \emph{The Eleventh International Conference on Learning Representations, {ICLR} 2023, Kigali, Rwanda, May 1-5, 2023}. OpenReview.net.

\bibitem[{Sun et~al.(2022)Sun, Yan, Abbeel, and Mordatch}]{sun2022quantifying}
Meiqi Sun, Wilson Yan, Pieter Abbeel, and Igor Mordatch. 2022.
\newblock \href {https://openreview.net/forum?id=LpBlkATV24M} {Quantifying uncertainty in foundation models via ensembles}.
\newblock In \emph{NeurIPS 2022 Workshop on Robustness in Sequence Modeling}.

\bibitem[{Sun et~al.(2023)Sun, Liu, Iter, Zhu, and Iyyer}]{sun2023does}
Simeng Sun, Yang Liu, Dan Iter, Chenguang Zhu, and Mohit Iyyer. 2023.
\newblock \href {https://doi.org/10.48550/arXiv.2302.11521} {How does in-context learning help prompt tuning?}
\newblock \emph{arXiv preprint arXiv:2302.11521}.

\bibitem[{Touvron et~al.(2023)Touvron, Lavril, Izacard, Martinet, Lachaux, Lacroix, Rozi{\`{e}}re, Goyal, Hambro, Azhar, Rodriguez, Joulin, Grave, and Lample}]{llama}
Hugo Touvron, Thibaut Lavril, Gautier Izacard, Xavier Martinet, Marie{-}Anne Lachaux, Timoth{\'{e}}e Lacroix, Baptiste Rozi{\`{e}}re, Naman Goyal, Eric Hambro, Faisal Azhar, Aur{\'{e}}lien Rodriguez, Armand Joulin, Edouard Grave, and Guillaume Lample. 2023.
\newblock \href {https://doi.org/10.48550/ARXIV.2302.13971} {Llama: Open and efficient foundation language models}.
\newblock \emph{CoRR}, abs/2302.13971.

\bibitem[{Wang et~al.(2019)Wang, Singh, Michael, Hill, Levy, and Bowman}]{wang2019glue}
Alex Wang, Amanpreet Singh, Julian Michael, Felix Hill, Omer Levy, and Samuel~R. Bowman. 2019.
\newblock \href {https://openreview.net/forum?id=rJ4km2R5t7} {{GLUE}: A multi-task benchmark and analysis platform for natural language understanding}.
\newblock In the Proceedings of ICLR.

\bibitem[{Wang et~al.(2023{\natexlab{a}})Wang, Li, Chen, Zhu, Lin, Cao, Liu, Liu, and Sui}]{wang2023large}
Peiyi Wang, Lei Li, Liang Chen, Dawei Zhu, Binghuai Lin, Yunbo Cao, Qi~Liu, Tianyu Liu, and Zhifang Sui. 2023{\natexlab{a}}.
\newblock \href {http://proceedings.mlr.press/v139/zhao21c.html} {Large language models are not fair evaluators}.
\newblock \emph{arXiv preprint arXiv:2305.17926}.

\bibitem[{Wang et~al.(2023{\natexlab{b}})Wang, Aitchison, and Rudolph}]{wang2023lora}
Xi~Wang, Laurence Aitchison, and Maja Rudolph. 2023{\natexlab{b}}.
\newblock \href {https://doi.org/10.48550/arXiv.2310.00035} {Lora ensembles for large language model fine-tuning}.
\newblock \emph{arXiv preprint arXiv:2310.00035}.

\bibitem[{Wei et~al.(2023)Wei, Hou, Lampinen, Chen, Huang, Tay, Chen, Lu, Zhou, Ma, and Le}]{wei2023symbol}
Jerry Wei, Le~Hou, Andrew Lampinen, Xiangning Chen, Da~Huang, Yi~Tay, Xinyun Chen, Yifeng Lu, Denny Zhou, Tengyu Ma, and Quoc Le. 2023.
\newblock \href {https://doi.org/10.18653/v1/2023.emnlp-main.61} {Symbol tuning improves in-context learning in language models}.
\newblock In \emph{Proceedings of the 2023 Conference on Empirical Methods in Natural Language Processing}, pages 968--979, Singapore. Association for Computational Linguistics.

\bibitem[{Wenzel et~al.(2020)Wenzel, Snoek, Tran, and Jenatton}]{wenzel2020hyperparameter}
Florian Wenzel, Jasper Snoek, Dustin Tran, and Rodolphe Jenatton. 2020.
\newblock \href {https://proceedings.neurips.cc/paper/2020/hash/481fbfa59da2581098e841b7afc122f1-Abstract.html} {Hyperparameter ensembles for robustness and uncertainty quantification}.
\newblock \emph{Advances in Neural Information Processing Systems}, 33:6514--6527.

\bibitem[{Yao et~al.(2023)Yao, Chen, Zou, Lu, Li, Zhang, Liu, Hendler, and Wang}]{yao2023more}
Bingsheng Yao, Guiming Chen, Ruishi Zou, Yuxuan Lu, Jiachen Li, Shao Zhang, Sijia Liu, James Hendler, and Dakuo Wang. 2023.
\newblock \href {https://doi.org/10.48550/arXiv.2311.09782} {More samples or more prompt inputs? exploring effective in-context sampling for llm few-shot prompt engineering}.
\newblock \emph{arXiv preprint arXiv:2311.09782}.

\bibitem[{Ye et~al.(2023)Ye, Hwang, Yang, Yun, Kim, and Seo}]{ye2023context}
Seonghyeon Ye, Hyeonbin Hwang, Sohee Yang, Hyeongu Yun, Yireun Kim, and Minjoon Seo. 2023.
\newblock \href {https://doi.org/10.48550/arXiv.2302.14691} {In-context instruction learning}.
\newblock \emph{arXiv preprint arXiv:2302.14691}.

\bibitem[{Zhang et~al.(2025)Zhang, Liu, and Patras}]{zhang-etal-2025-get}
Zhaohan Zhang, Ziquan Liu, and Ioannis Patras. 2025.
\newblock \href {https://aclanthology.org/2025.coling-main.726/} {Get confused cautiously: Textual sequence memorization erasure with selective entropy maximization}.
\newblock In \emph{Proceedings of the 31st International Conference on Computational Linguistics}, pages 10924--10939, Abu Dhabi, UAE. Association for Computational Linguistics.

\bibitem[{Zhao et~al.(2021)Zhao, Wallace, Feng, Klein, and Singh}]{zhao2021calibrate}
Zihao Zhao, Eric Wallace, Shi Feng, Dan Klein, and Sameer Singh. 2021.
\newblock \href {http://proceedings.mlr.press/v139/zhao21c.html} {Calibrate before use: Improving few-shot performance of language models}.
\newblock In \emph{International Conference on Machine Learning}, pages 12697--12706. PMLR.

\bibitem[{Zhou et~al.(2024{\natexlab{a}})Zhou, Wan, Proleev, Mincu, Chen, Heller, and Roy}]{zhou2023batch}
Han Zhou, Xingchen Wan, Lev Proleev, Diana Mincu, Jilin Chen, Katherine Heller, and Subhrajit Roy. 2024{\natexlab{a}}.
\newblock \href {https://openreview.net/forum?id=L3FHMoKZcS} {Batch calibration: Rethinking calibration for in-context learning and prompt engineering}.
\newblock In \emph{The Twelfth International Conference on Learning Representations}.

\bibitem[{Zhou et~al.(2023)Zhou, Wan, Vuli{\'c}, and Korhonen}]{zhou-etal-2023-survival}
Han Zhou, Xingchen Wan, Ivan Vuli{\'c}, and Anna Korhonen. 2023.
\newblock \href {https://aclanthology.org/2023.findings-emnlp.870} {Survival of the most influential prompts: Efficient black-box prompt search via clustering and pruning}.
\newblock In \emph{Findings of the Association for Computational Linguistics: EMNLP 2023}, pages 13064--13077, Singapore. Association for Computational Linguistics.

\bibitem[{Zhou et~al.(2024{\natexlab{b}})Zhou, Wan, Vuli{\'c}, and Korhonen}]{zhou2023autopeft}
Han Zhou, Xingchen Wan, Ivan Vuli{\'c}, and Anna Korhonen. 2024{\natexlab{b}}.
\newblock \href {https://doi.org/10.48550/arXiv.2301.12132} {Autopeft: Automatic configuration search for parameter-efficient fine-tuning}.
\newblock \emph{Transactions of the Association for Computational Linguistics}, 12.

\bibitem[{Zhu et~al.(2023)Zhu, Hao, He, Song, Zhang, Hu, Wei, Wang, and Lu}]{zhu2023clean}
Wenhong Zhu, Hongkun Hao, Zhiwei He, Yunze Song, Yumeng Zhang, Hanxu Hu, Yiran Wei, Rui Wang, and Hongyuan Lu. 2023.
\newblock \href {https://doi.org/10.48550/arXiv.2311.09154} {Clean-eval: Clean evaluation on contaminated large language models}.
\newblock \emph{arXiv preprint arXiv:2311.09154}.

\end{thebibliography}
